%% file: main.tex
\documentclass[acmsmall, nonacm]{acmart}

\usepackage{kotex}
\usepackage{comment}
\usepackage{subcaption}
\usepackage{bbm}

\AtBeginDocument{%
  \providecommand\BibTeX{{%
    \normalfont B\kern-0.5em{\scshape i\kern-0.25em b}\kern-0.8em\TeX}}}

\setcopyright{acmcopyright}

\acmJournal{TIST}
\acmMonth{9}

\begin{document}

\title{Explainable Product Classification for Customs}

\author{Eunji Lee}
\email{mk35471@gmail.com}
\orcid{0000-0002-1677-4102}
\affiliation{%
  \institution{School of Computing, KAIST}
  \streetaddress{291 Daehak-ro}
  \city{Daejeon}
  \country{Republic of Korea}
  \postcode{34141}
}

\author{Sihyeon Kim}
\affiliation{%
  \institution{School of Computing, KAIST}
  \city{Daejeon}
  \country{Republic of Korea}}
\email{sihk@kaist.ac.kr}

\author{Sundong Kim}
\authornote{Corresponding author: sundong@gist.ac.kr}
\affiliation{%
  \institution{AI Graduate School, GIST}
  \streetaddress{123 Cheomdangwagi-ro}
  \city{Gwangju}
  \country{Republic of Korea}}
\email{sundong@gist.ac.kr} 
\orcid{0000-0001-9687-2409}
  
\author{Soyeon Jung}
\affiliation{%
  \institution{ICT and Data Policy Bureau, Korea Customs Service}
  \city{Daejeon}
  \country{Republic of Korea}}
\email{jsy6519@korea.kr}  

\author{Heeja Kim}
\affiliation{%
  \institution{Customs Valuation and Classification Institute, Korea Customs Service}
  \city{Daejeon}
  \country{Republic of Korea}}
\email{tart75@korea.kr}  

\author{Meeyoung Cha}
\affiliation{%
  \institution{School of Computing, KAIST}
  \city{Daejeon}
  \country{Republic of Korea}}
\email{meeyoung.cha@kaist.ac.kr}
\orcid{0000-0003-4085-9648}

\renewcommand{\shortauthors}{Eunji, et al.}

\keywords{datasets, neural networks, gaze detection, text tagging}
\input{0_abstract}

\newcommand{\ej}[1]{\textcolor{orange}{#1}}
\newcommand{\sd}[1]{\textcolor{magenta}{#1}}
\newcommand{\sh}[1]{\textcolor{cyan}{#1}}
\maketitle

\input{1_introduction}

\input{2_related_work.tex}

\input{3_dataset}
\input{4_model}
\input{5_quantitative_study}
\input{6_qualitative_study}

\input{7_discussion}

\input{8_conclusion}
\begin{acks}
This research was supported by the Institute for Basic Science (IBS-R029-C2), NRF grant (RS-2023-00240062), and the IITP grant (RS-2023-00216011, 2019-0-01842) by the Ministry of Science and ICT in Korea. We thank Minsoo Song, Junok Kang, Sungdae Ji, Yeonsoo Choi from Korea Customs Service for their insightful discussions. 
\end{acks}

\bibliographystyle{ACM-Reference-Format}
\bibliography{references}











\end{document}

%% file: 0_abstract.tex
\begin{abstract}
The task of assigning internationally accepted commodity codes (aka HS codes) to traded goods is a critical function of customs offices. Like court decisions made by judges, this task follows the doctrine of precedent and can be nontrivial even for experienced officers. Together with the Korea Customs Service (KCS), we propose a first-ever explainable decision supporting model that suggests the most likely subheadings (i.e., the first six digits) of the HS code. The model also provides reasoning for its suggestion in the form of a document that is interpretable by customs officers. We evaluated the model using 5,000 cases that recently received a classification request. The results showed that the top-3 suggestions made by our model had an accuracy of 93.9\% when classifying 925 challenging subheadings. A user study with 32 customs experts further confirmed that our algorithmic suggestions accompanied by explainable reasonings, can substantially reduce the time and effort taken by customs officers for classification reviews. 
\end{abstract}



\ccsdesc[500]{Information systems~Expert systems}
\ccsdesc[500]{Computing methodologies~Natural language processing}
\ccsdesc[500]{Applied computing~E-government}

\keywords{Product classification, Interpretability, Decision support, Human-centered explainable AI}

%% file: 1_introduction.tex
\section{Introduction}
With the continuing advances in artificial intelligence, computational models are now being used to automate not only simple laborious tasks but also complex tasks that once seemed irreplaceable by machines. One example is self-driving cars, which are now available from a myriad of brands like Tesla and Google. AI is taking over the mundane task of steering the car, and rapidly learning to handle unknown scenarios. In legal sectors, tribunal and court decisions are being assisted by AI~\cite{sourdin2018judge}. Many other domains are adopting AI in their core functions, including medical decisions, surveillance, climate modeling, and financial predictions.

However, it remains unclear whether AI can completely replace human tasks. In particular, some argue that AI shouldn't be a final arbiter for mission-critical tasks that require human reasoning~\cite{mckay2020predicting}. For example, while court decisions must be based on an in-depth understanding of the precedent and relevant laws; they are also subject to subtlety and sometimes need moral and policy judgments. Others have argued that, for this reason, outcomes of data-driven models cannot be perceived as objective by the public nor replace field experts~\cite{de2021perils}.

\begin{figure}[t!]
\centerline{
      \includegraphics[width=\linewidth]{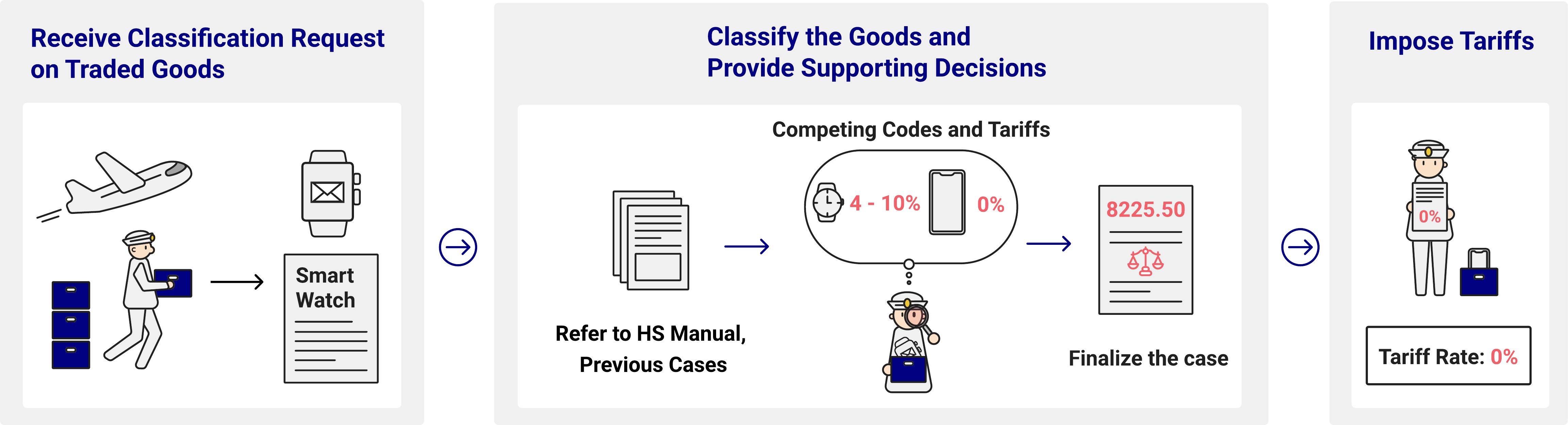}}
      \caption{Overview of the HS code classification procedure at the customs office. 
       } 
\label{fig:overview}
\end{figure}

Explainable AI (XAI) is a promising alternative to conventional AI, for use as an assistance tool in various sectors. XAI offers human interpretable reasoning with algorithmic suggestions and can assist humans in taking an arbiter role. At the same time, it can substantially reduce the time and effort needed for humans to complete complex tasks. In many sectors, when implementing AI in high-stakes scenarios, XAI is considered a critical requirement~\cite{langer2021we}.

This paper presents a first-ever XAI model to assist customs officers in assigning commodity codes (aka HS codes) as they classify traded goods. As in the legal sectors, general classification guidelines are written in an HS Classification Handbook\footnote{https://www.wcoomdpublications.org/en/products/harmonized-system/explanatory-notes-2022}, which is internationally accepted. However, classification is a nontrivial task even for experienced customs officers. HS code determines the tariff borne by the importer, and hence some decisions can lead to an international dispute. Disputed cases are handled by national and international customs committees, whose decisions are later used as the doctrine of precedent. Our goal in this work is to implement an AI model that provides top suggestions along with human-interpretable reasoning.

Figure~\ref{fig:overview} shows an overview of the HS code classification procedure. Traded goods need to be declared with HS codes based on the HS manual, and the general regulations for HS. This decision is sensitive as it determines the tariff rate, yet it is not obvious, as similar goods may need to be assigned different HS codes and vice versa. Customs experts examine disputed cases, and their decisions are logged at customs offices. The primary goal of our AI model is to assist this decision process by giving algorithmic suggestions along with human interpretable reasoning.

Our AI model operates in two stages. First, it predicts item classification based on the text description of the goods. Second, it retrieves evidence about each candidate from the HS manual. As a result, the AI suggestion consists of candidate codes and the relevant key sentences in the HS manual as explainable evidence. The AI model runs on top of the latest Natural Language Processing (NLP) model and has been tested on real HS classification cases involving mechanical and electrical equipment (as listed in Chapter 84, 85, and 90), the goods categories that are known to be most challenging for human officers because of their similarity. Our model showed a high accuracy of 93.9\% when the top-3 candidates of the 6-digit HS codes were suggested for 925 classes. 

\textcolor {black} {This work greatly benefits from our collaborating partners at the Korea Customs Service (KCS), from which we recruited 32 field officers to help test the efficacy of the prototype AI model. Customs field officers of varying career experience participated in our usability survey. The survey data indicated that the proposed AI model was perceived to be `helpful' (by 85\% of the respondents) as a supporting tool, and officers found value in its ability to reduce the time needed for screening candidate codes. In particular, the AI model was perceived to be more helpful by officers with shorter field experience, who valued receiving multiple candidate suggestions and their interpretable evidence. These findings suggest that AI models can be used to train and assist novice customs officers and contribute to cost reduction. Currently, we are beta-testing the HS classification service for target users at Korea Customs Service: see \url{https://ds.ibs.re.kr/product-classification/}. It is worth noting that AI is increasingly providing support in various domains, such as legal cases with predictive judgment systems~\cite{santosh-etal-2022-deconfounding} and search systems\footnote{https://deepjudge.ai/}. Similarly, this work has the potential to be beneficial in the context of HS code classification tasks. In conclusion, we discuss the implications of these findings and outline potential future directions. We conclude with a discussion of implications and future directions.}

%% file: 2_related_work.tex
\section{Preliminaries}

\subsection{HS Classification for Traded Products}

According to the World Customs Organization (WCO), the number of import and export declarations worldwide reached 500 million in 2020. Events like the outbreak of coronavirus disease 2019 (COVID-19) have led to a surge in the cross-national imports of e-commerce goods, where for instance, Korea accounted for 63.5 million in 2020, a 48\% increase compared to the previous year~\cite{e-commerce-stats}. 
As global transactions increase and traded products become diversified, the management of standards for categorizing numerous products---i.e., Harmonized Commodity Description and Coding System, or Harmonized System (HS) for short--- is becoming crucial. The HS is an international standard for classifying goods. From live animals to electronic devices, each product is classified under one of 5,387 subheadings (the first six digits of the HS codes) that meet international conventions~\cite{HS_compendium}.
This code determines critical trade decisions like tariff rates and import and export requirements. 

HS code classification is nontrivial and requires a high degree of expertise since it determines the tariff rate. Securing tariffs is vital for fiscal income in many countries. The share of tax revenue secured through customs offices is nearly 20\% worldwide and exceeds 40\% in West African countries.\footnote{WCO annual report shows the proportion of revenue collected by customs in tax revenue of each country (pp.~46-91): \url{https://tinyurl.com/yxjvn9mz}} In addition, tariff rates are directly linked to the price of goods, affecting their global competitiveness. Therefore, both importers and exporters pay special attention to product declarations. Customs authorities scrutinize the submitted HS codes of declared goods and correct them if needed. Simple errors can be corrected by amending the declaration or sending a request for correction. If customs administrations find evidence of smuggling or deliberate false declarations for tax evasion purposes, then importers are punished by customs acts. 

The process of classifying a product is complex because human interpretations may not always be consistent, which can lead to an international dispute when a ruling between customs authorities differs or differs between the companies and customs authorities. For example, when smartwatches were first released, tariffs varied across importing countries due to the absence of a classification standard. As shown in Figure~\ref{fig:overview}, tariff rates for wireless communication devices are 0\%, but 4--10\% for watches, which led to a dispute that was finally resolved by the WCO HS Committee in 2014. The committee classified smartwatches as wireless communication devices, and the manufacturer was able to save approximately \$13 million per year~\cite{smartwatch}. Furthermore, difficulties arise when the product has multiple characteristics or new characteristics that are not mentioned clearly in the guidelines. The customs administration operates a pre-examination system, allowing import and export companies to request customs to review their items before formal declarations. Korea Customs Service receives approximately 6,000 applications for pre-examination every year. With the increasing complexity of goods, the processing time has increased from 20.4 to 25.8 days per inspection since 2018. The main reason for this is the detailed review process since the emergence of the HS code and that the corresponding tax rate can differ, even for similar-looking items. For example, tariff rates for television (HS 8528.59) are 8\% but 0\% for PC monitors (HS 8528.52).

The HS code classification process includes a review of the descriptions submitted by applicants and relevant cases in the past. Experts adjust to the HS manual that includes Explanatory notes of the HS~\cite{HS_manual} for standard code descriptions, General Rules of Interpretation (GRI)~\cite{GRI} for decision-making criteria, HS Nomenclature, and HS Compendium of Classification Opinions. We designed our model to reflect this process. First, it suggests the first six digits of HS codes (called \textit{subheadings}) based on product descriptions with pretrained language models. Then, it retrieves key sentences from the HS manual that are most related to the product. The retrieved sentences act as supporting facts to support the final decision statement by the officer, which will be provided to the importers and exporters who request classification.

\subsection{HS Code and Its Classification}

The WCO explains that the Harmonized Commodity Description and Coding System (popularly known as the Harmonized System or the HS) is one of the most successful instruments ever developed by the WCO. It is a multipurpose goods nomenclature used by more than 200 countries and Customs or Economic Unions as the basis for Customs tariffs and the compilation of international trade statistics.\footnote{\url{http://www.wcoomd.org/en/topics/nomenclature/overview/what-is-the-harmonized-system.aspx}}  

All the items that go through customs are assigned an HS code, an internationally standardized system of names and numbers to classify traded products to determine tariffs. As an internationally recognized standard, the first six digits of the HS code (HS6) are the same for all countries. Countries have added more digits to their respective HS code systems for further classification. HS6 includes the following three components: 

\begin{enumerate}
\item  \textbf{Chapter}: the first two digits of the HS code, which contains 96 categories from 01 to 99. 
\item  \textbf{Heading}: the first four digits of the HS code, groups similar characteristics of goods within a chapter.  
\item  \textbf{Subheading}: the first six digits the of HS code, groups goods within a heading. 
\end{enumerate}

\begin{figure*}[htpb!]
    \centering
    \includegraphics[width=0.8\linewidth]{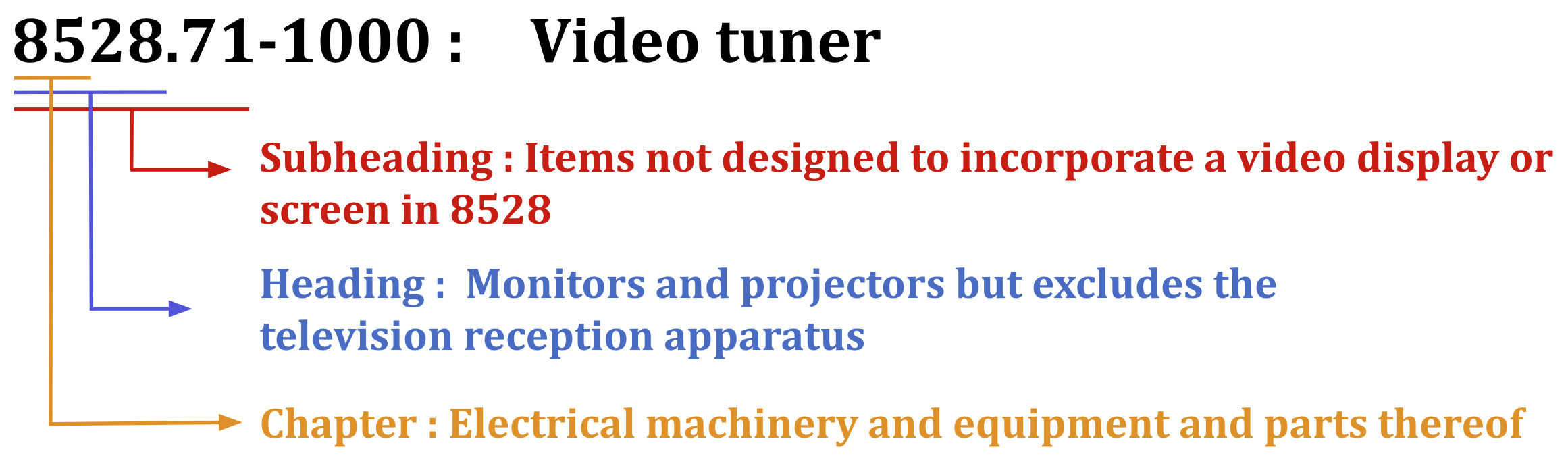}
      \caption{Hierarchical structure of the HS code.} 
\label{fig:data_imbalance}
\end{figure*}

As a characteristic data property, we note that HS code heading and subheading data are skewed in terms of their frequency. Popular headings and subheadings appear disproportionately more times in the classification data, which could help the model learn the data traits. Figure~\ref{fig:data_imbalance} shows an example of the frequent and infrequent headings in the Chapter 85 data, where the first three headings appear over 8000 times each. The least popular item appears less than 500 times. Such popularity also has evolving temporal trends, with some headings becoming less requested (or more requested) over time. We later discuss how the skewed popularity and temporal dynamics affect quality in the Discussion section. 

\begin{figure*}[htpb!]
    \centering
    \includegraphics[width=0.49\linewidth]{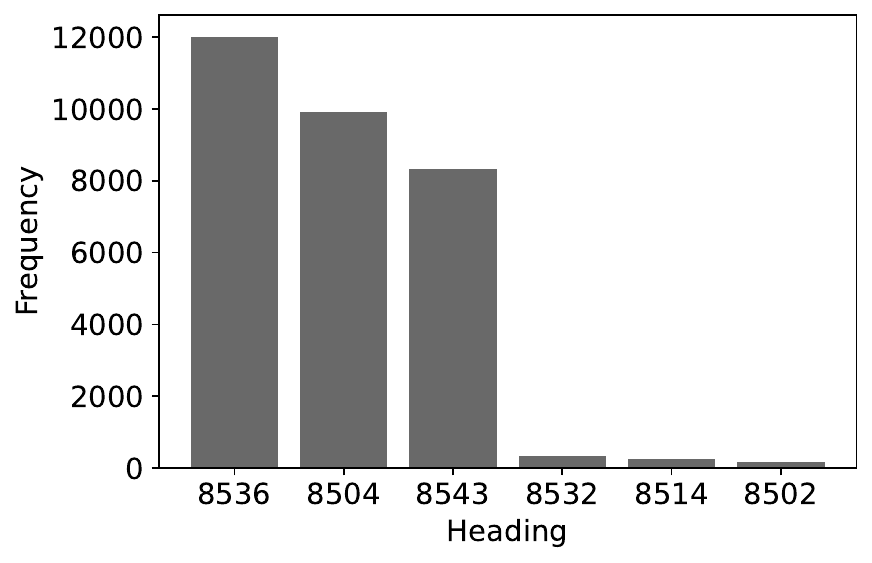}
      \caption{Imbalanced heading data in decision case.} 
\label{fig:data_imbalance}
\end{figure*}

Recent studies have utilized machine learning approaches to classify HS codes using text descriptions of the declared goods. These approaches include the $k$-nearest neighbor, support vector machine (SVM), Adaboost~\cite{DING20151462}, and neural networks~\cite{NNTHS2021}. For text embedding, BERT~\cite{kenton2019bert} and ELECTRA~\cite{clark2020electra} have been introduced. To capture semantic information, studies have used neural machine translators~\cite{10.1007/978-3-030-44322-1_22} and other transformer-based algorithms~\cite{luppes2019classifying}. Some have utilized hierarchical relationships between the HS codes and the co-occurrence of the words using background nets~\cite{Customs2019}. Similar studies have understood short texts and classified them into a larger hierarchy using class taxonomy~\cite{shen2021taxoclass}, metadata~\cite{zhang2021match}, and hyperbolic embedding~\cite{chen2020hyperim}, which can be applied to HS prediction. Some studies have also utilized image information for classification~\cite{lee2020cnn}. 
However, most approaches have focused on the classification itself, and lack any explanation.

\subsection{Interpretable Text Classification} 
A large number of studies have classified text datasets using AI models. Recent studies have added interpretability to the classification model to solve the black-box nature of the deep learning model. Commonly, the classification model highlights the essential parts of a given sample to show where the model had concentrated. Some researchers provide an explanation at the sample data level and the general behavior of the model, showing frequent words and queries in learning~\cite{wang2020icapsnets}. Self-interpretable convolutional neural networks~\cite{zhao2021self} have been suggested, and approaches to use max-pooling have helped interpret predictions with input tokens~\cite{cheng2019interpretable}. Other studies have utilized relations among text to extract linguistic features and to understand how a language model works~\cite{dasgupta2018automatic}. Despite the ongoing efforts on interpretable text classification, an XAI model for HS code classification is unknown. Deciding HS code requires deep reasoning and it sensitively affects multiple parties; the whole process can benefit from explainable models with strong retrieval capability.

Sentence retrieval is a key module that provides explanations for the question and answering (QA) problems~\cite{thayaparan-etal-2019-identifying, wang2019evidence}. Explanations are critical for QA tasks, where leading researchers have built numerous datasets with annotated sentences such as HotpotQA~\cite{yang-etal-2018-hotpotqa} and QASC~\cite{Khot2020QASCAD}. Various retrieval techniques have been introduced with these datasets, such as self-attention~\cite{wang2017gated}, bi-attention~\cite{seo2017bidirectional}, graph-based network~\cite{fang2020hierarchical}, and unification-based approach~\cite{valentino2020unification}. However, extending these approaches is difficult when handling datasets in which ground-truth supporting facts are not given. In an unsupervised setting, alignment-based methods based on word usage and similarity have been used, such as term frequency-inverse document frequency (TF-IDF)~\cite{ramos2003using} and word-mover distance~\cite{kim2017bridging}. The latter identifies related words when no direct matches are found between a query and a document. As the language model matures, a similarity-based pipeline can arrive at answers and supporting sentences with high performance~\cite{groeneveld2020simple}, shedding light on real-world applications with limited annotation. In the context of our research, we formulated our problem as semi-supervised sentence retrieval. We leveraged the domain expertise contained within a subset of our dataset to replicate the decision-making process of experts and enhance a model designed to assist human users. This model capitalizes on the knowledge contributed by expert annotators and builds upon the groundwork established by the previously mentioned methods.

%% file: 3_dataset.tex
\section{Datasets}



We obtained 20 years of recent goods classification data from the Korea Customs Valuation and Classification Institute.\footnote{\url{https://www.customs.go.kr/cvnci/main.do}} Our data span three chapters: Chapter 85 (for electrical equipments), 84 (for mechanical appliances), and 90 (for optical and photographic instruments). 
These chapters are known as the most challenging to classify by experts~\cite{lee2021kaia}, because goods in these chapters have multiple functions and do not easily fit into a single HS category~\cite{park2019fallacy}. As a result, Chapter 85-related goods receive the most requests for expert review, accounting for nearly 17\% of all classification requests in 2020 followed by Chapter 84 the second (10.1\%), and Chapter 90 the fourth (6\%) by Korea Customs Valuation and Classification Institute.

Furthermore, goods in these three categories share similar descriptions.\footnote{For example, mixing units used in the sound recording are classified into heading 8543, but if it's specialized for cinematography, it belongs to heading 9010.}  The task is a multi-class classification problem since these three chapters contain 163 headings and 925 subheadings in total. According to the Institute, the average classification review period to resolve classification requests in Chapter 85 is nearly 37.2 days, 36.3 days for Chapter 84, and 33.3 days for Chapter 90, far longer than the average time required for resolving goods in other categories (taking 25.9 days on average).

\input{tables/data_distribution}

Table~\ref{table:data_distribution} shows the data summary we utilized in our study. The Korean data, totaling 17,068 cases, are in three levels of difficulty: 16,035 cases (or 93.9\%) were resolved by the field officers at the Institute. Not all cases can be resolved by field officers, and some are moved up to be resolved at the HS council accounting for 634 (or 3.7\%) of the studied cases. The most challenging cases that remain unresolved at this level are escalated to the HS committee to receive a final decision, accounting for 399 (or 2.3\%) of the studied cases. We obtained the detailed supporting facts and decisions for these 1,033 contentious cases and used such information for training data.

In addition to the Korean data, International cases from 50 countries were included. The HS classification is a six-digit standard, called a subheading, for classifying globally traded products. The Institute provided international cases to use as additional training resources. We translated them into one language (i.e., Korean) and used them together to train the model. We used the first six-digits of the HS code (i.e., subheading) as it is internationally standardized. Some of the received Korean International data are made public under the international HS directory on the Customs Law Information Portal website.\footnote{\url{https://unipass.customs.go.kr/clip/index.do}} 

\begin{figure}[t!]

       \begin{subfigure}{.49\textwidth}
        \centering
        \includegraphics[width = \linewidth]{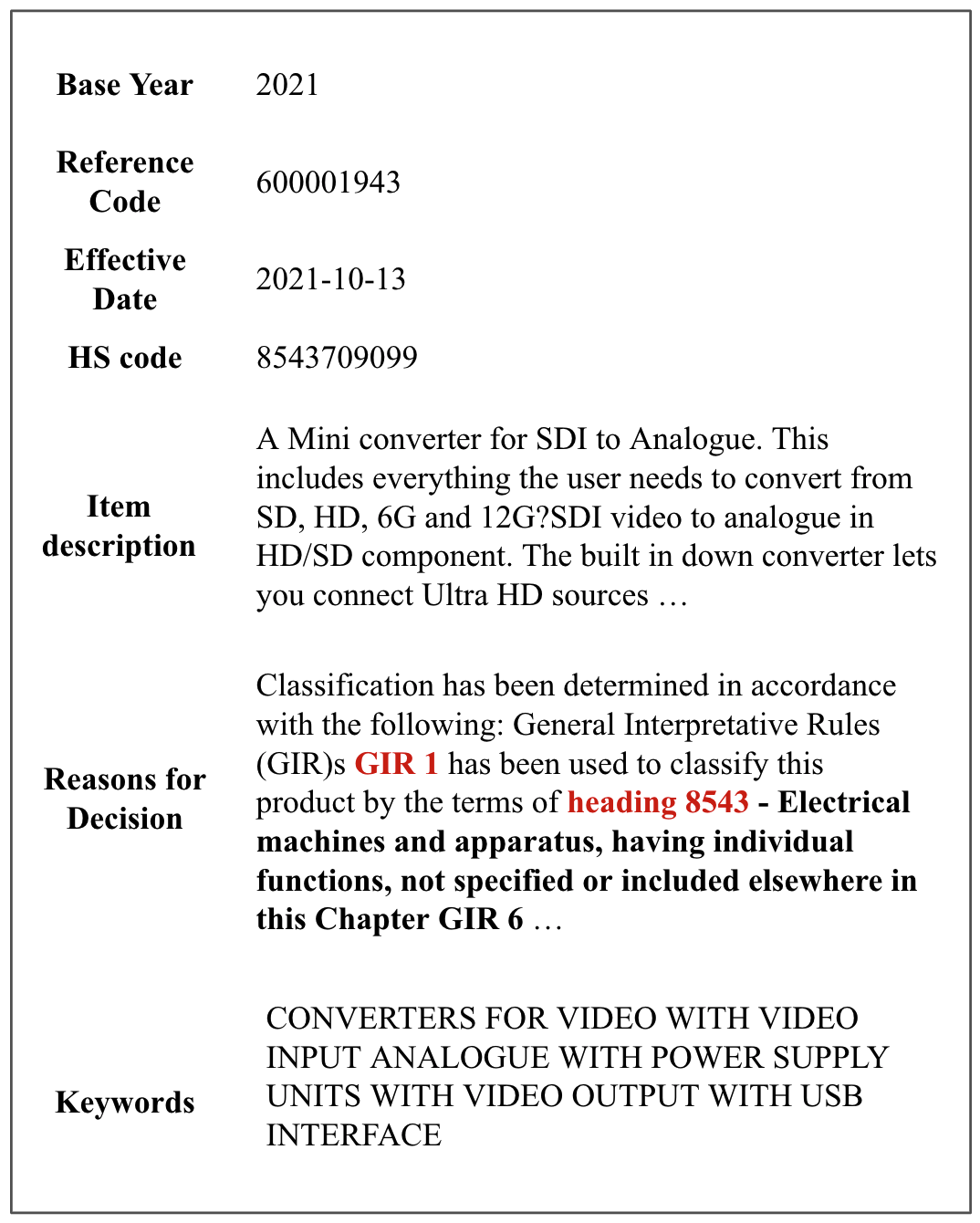}
        \caption{Decision case sample}
        \end{subfigure}\hfill
       \begin{subfigure}{.496\textwidth}
        \centering
        \includegraphics[width = \linewidth]{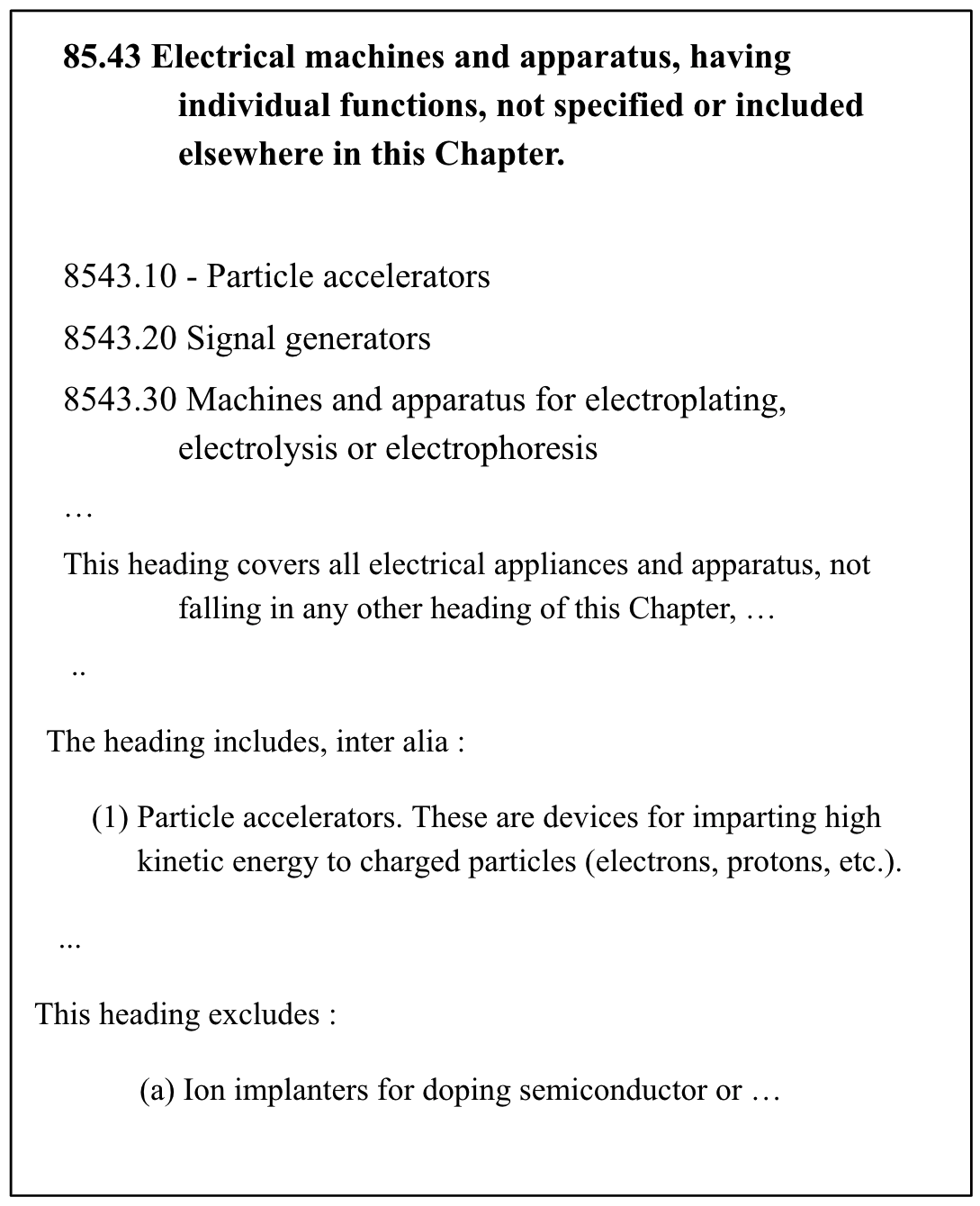}
        \caption{HS manual sample}
        \end{subfigure}\hfill
      \caption{Sample decision data and the relevant HS manual. In (a), the given item is classified as 8543709099, `Others in Other machines and apparatus' by H.M. Revenue and Customs. Base year, reference code, effective date, HS code, item description, reasons for the decision, and keywords are given in this example case. In (b), the manual provides the characteristics and standards of each heading (8543 here) in detail, and it includes a one-liner description of every subheading.
      } 
\label{fig:example}
\end{figure}

Figure~\ref{fig:example} (a) shows a sample classification decision from the studied data, along with the matching HS manual. The GIR in the figure represents the General Interpretative Rules of the HS, consisting of six principles. The first principle (i.e., GIR 1) states that classification shall be determined according to the terms of the headings and any relative section or chapter notes. This example shows that the decision for HS code 854370XXXX was assigned because the description of the goods matched the guidelines for goods in heading 8543. Note, that our goal is to build an AI model that suggests the top 6 digits of the HS code (i.e., up to the subheading level).\footnote{The remaining digits describe information such as the color, shape, and material of the goods and could be determined by inspection easily.}

As mentioned earlier, the HS codes comprise 1,224 headings within 97 Chapters, arranged in 21 sections of the manual. Figure~\ref{fig:example} (b) shows an example heading level from the HS manual, which starts from a heading description and is followed by a subheading. The HS manual also includes an explanation of the heading and important terminologies. In addition, it gives a list of items the heading includes and excludes.

%% file: tables/data_distribution.tex



\begin{table}[htpb!]
\caption{The number of cases used for item classification. Three chapters---which are often confused during classification---were used for this study. The dataset includes contentious cases that were initially withheld and later were resolved by the HS council and HS committee.}
\label{table:data_distribution}
\centering
\small
\begin{tabular}{c|ccc|c}
\hline
        &         & Korean case data  &         & International  \\ \cline{2-4}
Chapter & General & Council & Committee   & case data                    \\ \hline
84      & 5,115   &  231     & 122       & 55,966              \\
85      & 6,434   & 237     & 192          &  122,221             \\
90      & 4,486   & 166     & 85         & 31,448              \\ \hline
\end{tabular}
\end{table}

%% file: 4_model.tex
\begin{figure*}[t]
\centerline{
      \includegraphics[width=.95\linewidth]{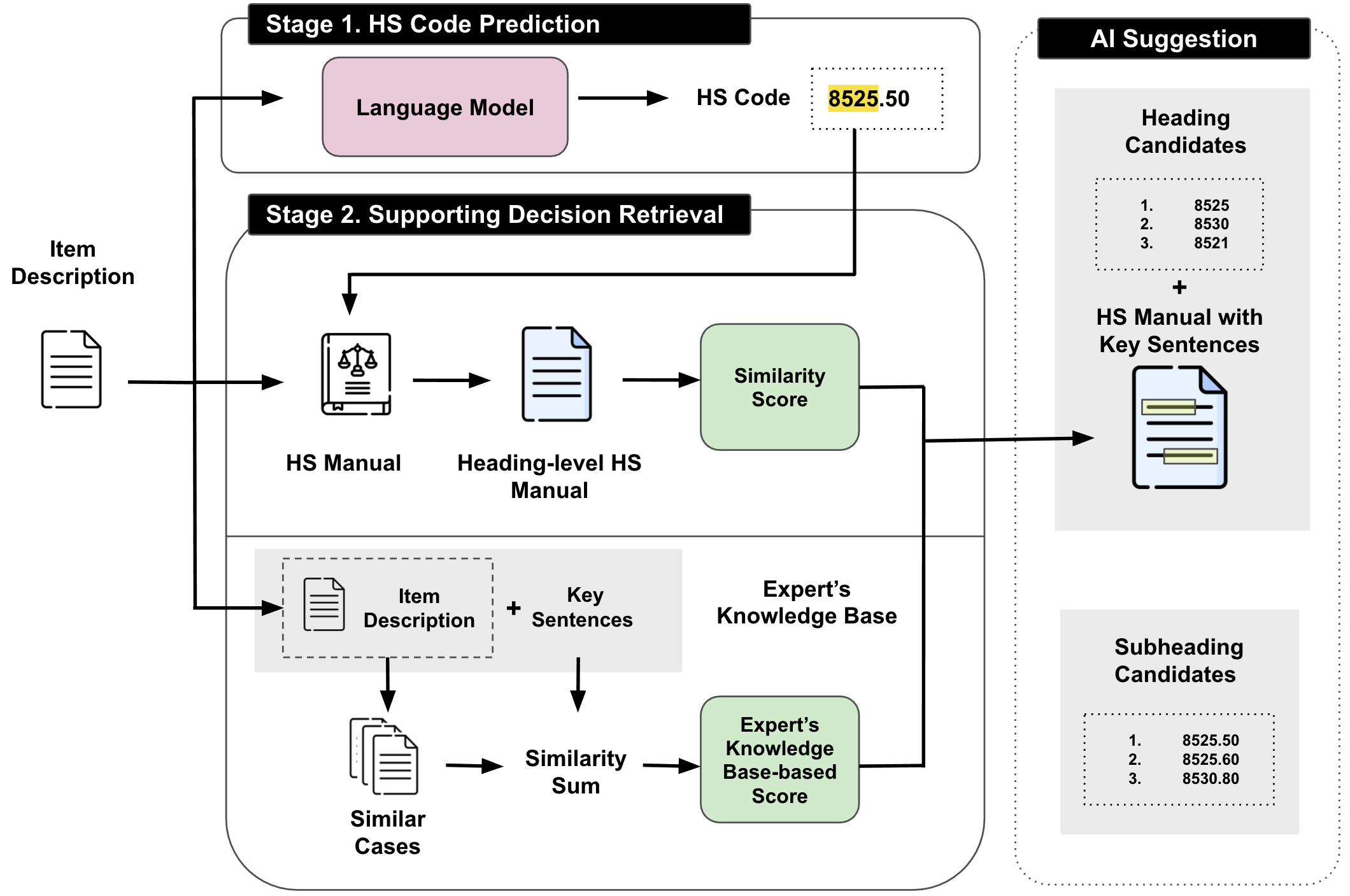}}
      \caption{The illustration of the proposed HS code classification supporting model. Stage 1 uses a language model and predicts the HS code of the goods. Stage 2 retrieves key sentences as supporting evidence using two measures: similarity (text similarity with HS manual) and expert knowledge (similar cases from the precedent). Top-3 suggestions at the subheading level (i.e., first six digits) are given in this example.
      } 
\label{fig:model}
\end{figure*}

\section{Interpretable Product Classification Model}

We now present an explainable product classification model. Our model takes the goods description as input and suggests the appropriate subheading (HS6 or first six digits) candidates along with some evidence. As evidence, we will show relevant sentences in the HS manual that could support the decision. Figure~\ref{fig:model} illustrates the flow of the model, which is divided into two stages: HS code prediction and supporting decision retrieval. We describe each step in detail.

\subsection{Stage 1 : HS Code Prediction}

Stage 1 uses a language model and predicts the HS code of the goods. Let $\mathcal{D} = \{D_1,\cdots, D_N\}$ be a collection of decision cases, where each case $D_i \in \mathcal{D}$ is a pair of the item description $\mathbf{x}_{i}$ and its one-hot encoded heading label $\mathbf{y}_{i}$. After translating all of the goods description into a common language (\textit{i.e., Korean, for example}), we use a language model as a description encoder, $e_{\theta}$, to map a sequence of words $\mathbf{x}_{i}$ into embedding space $\mathbb{R}^d$. Item embedding $e_{\theta}(\mathbf{x}_{i})$ goes through the classification head, and the model is trained to minimize the loss $\mathcal{L}$ between true probability $\mathbf{y}_{i}$ and predicted probability $\mathbf{\hat{y}}_{i} = e_{\theta}(\mathbf{x}_{i})\cdot W$, where $W \in \mathbb{R}^{d\times\dim(\textbf{y}_i)}$ is a trainable weight matrix of the classification head. Following the rule of assigning a single HS code to each product, we regard the problem as a multiclass classification and minimize the categorical cross-entropy loss $H$.   
\begin{align} 
    \mathcal{L} = -\frac{1}{|\mathcal{D}|}\sum_{{\mathbf{x}_i, \mathbf{y}_i}\in\mathcal{D}}  H(\mathbf{y}_i, \mathbf{\hat{y}}_i)
    \label{eq:heading_prediction}
\end{align}

\subsection{Stage 2 : Supporting Decision Retrieval}
  
Stage 2 then retrieves key sentences as supporting evidence. To identify informative sentences, we consider two measures. One is the text similarity between the goods description and the HS manual. Second is the knowledge accumulated by experts based on previous decisions. We describe these ideas more formally below.  

Let $\mathcal{M} = \{M_1,\cdots, M_K\}$ be a collection of sentences in a heading-level HS manual, find a set of relevant sentences $S_i = \{{M_j}\}$ with given goods description $\mathbf{x}_i$. To measure the relevancy between $\mathbf{x}_i$ and the HS manual sentence $M_k$, we define a relevance score $s(\mathbf{x}_i, M_k)$ as follows:
\begin{align}
    s(\mathbf{x}_i, M_k) =  s_s(\mathbf{x}_i, M_k) + \lambda s_e(\mathbf{x}_i, M_k),   
\end{align}
where $s_s(\mathbf{x}_i, M_k)$ is the text similarity score between two inputs, and $s_e(\mathbf{x}_i, M_k)$ is a score referring to sentences (supporting facts) written by experts to classify previous cases. $\lambda$ regulates the contribution between two values. The sentences with the highest score $s(\mathbf{x}_i, M_k)$ forms $S_i$. The following sections explain each score in detail.\\

\noindent $\bullet$ \textbf{Text Similarity. \quad}
The first score $s_s(\mathbf{x}_i, M_k)$ is the text similarity between the sentence $M_k$ and the given item description $\mathbf{x}_i$. Inspired by the AIR~\cite{yadav2020air} model, the text similarity measures the alignment between words in $M_k$ and $\mathbf{x}_i$. The score is high if the item description contains more words specialized in the category. $s_s(\mathbf{x}_i, M_k)$ is defined as follows:
\begin{align} 
    & s_s(\mathbf{x}_{i},M_k) = \sum_{l=1}^{|\mathbf{x}_{i}|} idf(d_l) \cdot align(d_l, M_k), \\
    & align(d_l, M_k) = \max_{t=1}^{|M_k|} \textit{CosSim}(d_l, m_t),
\end{align}
where $d_l$ and $m_t$ are the $l^{th}$ and $t^{th}$ terms of $\mathbf{x}_{i}$ and $M_k$, respectively. The cosine similarity ($CosSim$) is derived by embeddings of the two inputs from the trained language model in Stage 1, and $idf(d_l)$ is the inverse document frequency (IDF) of the word $d_l$.  \\

\noindent $\bullet$ \textbf{Expert Knowledge. \quad} \label{sec:kb}
The second score $s_e(\mathbf{x}_i, M_k)$ involves selecting key sentences based on past decisions. In certain intricate cases, pertinent information labeled as `Reasons for Decision' was extracted from experts' input, as depicted in Figure 4. These records pertained to the resolution of contentious cases undertaken by the HS Committee and HS Council. We formed a knowledge base $\mathcal{KB}=\{(\mathbf{x}^{\mathcal{KB}}_1, E^{\mathcal{KB}}_1),\cdots, (\mathbf{x}^{\mathcal{KB}}_m, E^{\mathcal{KB}}_m)\}$ by aggregating each contentious case $\mathbf{x}^{\mathcal{KB}}_j$ and its supporting facts $E^{\mathcal{KB}}_j = \{M_1, \cdots, M_a\}$. $m$ is the number of cases in the knowledge base $\mathcal{KB}$ and $a$ is the number of sentences for the $j$-th case. For instance, in the context of a classification case illustrated in Figure 4, the item description assumes the role of $\mathbf{x}_j$, and each quoted sentence, exemplified by phrases such as `heading 8543-Electrical machines and apparatus, having individual functions, not specified or included elsewhere in this Chapter.' is represented as $M_k$ in $E^{\mathcal{KB}}_j$. First, we pick the most relevant $k_{case}$ cases $S^{\mathcal{KB}}(\mathbf{x}_i)$ with a given item description $\mathbf{x}_i$. 
\begin{align}
    & S^{\mathcal{KB}}(\mathbf{x}_i) = TopK_{x^{\mathcal{KB}}_j \in \mathcal{KB}}(CosSim(\mathbf{x}_i, \mathbf{x}^{\mathcal{KB}}_j), k_{case}),
\end{align}
where $CosSim$ is the cosine similarity between two input embeddings and $TopK$ returns the most relevant $k_{case}$ pairs in $\mathcal{KB}$ with high similarity values. Once $S^{\mathcal{KB}}(\mathbf{x}_i)$ is decided, the $\mathcal{KB}$-based similarity score $s_e(\mathbf{x}_i, M_k)$ is defined as follows:
\begin{align}
    s_e(\mathbf{x}_i, M_k)  = \sum_{(x^{\mathcal{KB}}_j, E^{\mathcal{KB}}_j) \in S^{\mathcal{KB}}(\mathbf{x}_i)} {CosSim(\mathbf{x}_i, \mathbf{x}^{\mathcal{KB}}_j)\mathbbm{1}_{M_k \in E^{\mathcal{KB}}_j}}. 
\end{align}
This score represents the summation of cosine similarity values between the provided input $x_i$ and $x^{\mathcal{KB}}_j$ within the knowledge base, contingent upon the presence of $M_k$ in the knowledge base. For instance, if the sentence `heading 8543-Electrical machines \dots in this Chapter,' is quoted by experts in the top-k most similar previous cases, it will accrue a higher score in this context.


%% file: 5_quantitative_study.tex
\section{Quantitative Evaluation}

We tested the feasibility of the proposed AI model using extensive quantitative and qualitative experiments. The quantitative evaluation focused on model accuracy compared to alternative methods and the quality of the suggestions.

\subsection{Experimental Setting}
 
We split the data into non-overlapping training, testing, and validation sets. In doing so, we tried to preserve the time order of the data. The first 201,435 cases were used for training the model. The most recent 5,000 cases were used for testing (including 4,324 international and 676 Korean cases). The next latest 5,000 cases were used for the validation set (4,739 international cases and 261 Korean cases) for hyperparameter tuning.

Since the AI model is intended to be used as a supporting tool, we gave the participating human inspectors multiple suggestions to choose from. Algorithmic suggestions were given for the top-$k$ choices by accuracy for $k = 1, 3, 5$. In the retrieved sentence case study analysis, we measured recall and precision to evaluate the quality of the supporting factors.  

For the baseline, we used a long short-term memory (LSTM)-based model. The LSTM-based model was a winning model in the product categorization competition in Daum shopping,\footnote{\url{https://github.com/lime-robot/product-categories-classification}} which has a similar setting to our problem: predict the detailed category of e-commerce products using their descriptions. The model utilizes LSTM networks to obtain embedding from tokenized input texts.

Our model was implemented over three backbone language models: KoBERT~\cite{kenton2019bert}, KoELECTRA~\cite{clark2020electra}, and KLUE-RoBERTa~\cite{park2021klue} for the experiment. We used open-sourced implementations of KoBERT-base\footnote{\url{https://github.com/SKTBrain/KoBERT}}, KoELECTRA-base\footnote{\url{https://github.com/monologg/KoELECTRA}}, and KLUE-RoBERTa-base\footnote{\url{https://github.com/KLUE-benchmark/KLUE}}. The language models were trained for 100 epochs and evaluated when the validation accuracy was the highest. The embedding size of the language model was set to 768. We set the contribution regulating parameter $\lambda$ (Equation (3)) to 0.3, and the number of similar cases $k_{case}$ (Equation (7)) to 10 for sentence retrieval. The language model training took 40 hours and data preparation for sentence retrieval took 50 hours on an NVIDIA TITAN Xp. Inference and retrieval took less than 30 seconds.
\input{tables/main_results}

\subsection{Heading and Subheading Accuracy}

Table~\ref{table:main_results} shows the top-$k$ accuracy of the LSTM baseline and our model with different language models. Top-$k$ accuracy was tested in both heading (HS4, first four digits) and subheading (HS6, first six digits). Our model with the KLUE-RoBERTa backbone network suggests top-3 candidates with 95.5\% accuracy when classifying 163 headings, and 93.98\% accuracy for 925 subheadings. These results show that our AI model performed better than the LSTM-based winning model. The table also shows that our model will improve when a more powerful backbone network is used. Note that KoBERT uses 92M parameters, taking nearly 30 minutes to train one epoch. KoELECTRA uses 113M parameters and took 24 minutes for training. KLUE-RoBERTa uses 336M parameters and takes 70 minutes to train each epoch.

\input{tables/data_difficulty}

In addition, we measured the performance of contentious cases in the test dataset that the HS Committee and HS Council resolved. The KLUE-RoBERTa model was used for this experiment. Table~\ref{table:data_difficulty} shows that like human experts the AI model also had difficulty predicting the correct answer (i.e., a substantial drop in top-1 accuracy). Still, it provides helpful information with reasonably high top-3 and top-5 accuracy.

\input{tables/supporting_facts}
\subsection{Retrieved Key Sentences}

We mimicked the existing consulting documents in generating the supporting evidence for each suggestion. When field officers generate consulting documents, they quote sentences from the HS manual that give strong support for their ultimate decision. Because this document was generated manually, it had a common structure yet varied in content. Some documents also mentioned competing HS codes that were considered in the decision. \\

To evaluate the quality of the automatic evidence that our AI model generated, we took small samples of the consulting documents and compared them with the AI-generated ones. We compared quoted sentences from experts and retrieved sentences from our algorithm at the sentence level.

Table~\ref{tab:supporting_facts} shows an example of the evidence generated by the field officer (on top) and by the algorithm (bottom). This particular case of heading `8472' had 75 original sentences in the HS manual. We compared which sentences were highlighted by the algorithm and which ones were chosen as evidence by human experts. This test had a recall and precision of 0.75 each. The first sentence retrieved from our model matches the second sentence from the expert. Similarly, the second sentence from our model matches the third one from the expert, and the third sentence matches the fourth one from the expert. 

To test the quality of the evidential sentences, we also obtained 15 new contentious cases listed on the Customs Law Information Portal website. None of these cases had been used for training our model. Our AI model generated suggestions for these complex cases and retrieved seven sentences for these cases, based on feedback from the field officers, many of whom indicated seven as the most preferred number of sentences to view. \\
We report the recall scores between the automatically highlighted sentences and those indicated by a human expert. This evaluation focuses on the recall value because the AI model should not miss any key evidential sentences that human experts consider important. The precision metric is excluded since it can differ depending on the number of sentences to view. The average recall was 0.69 for the 15 cases, indicating that the AI model chose the same evidential sentence as humans with a probability of 69\%.

\begin{figure*}[htpb!]
    \begin{subfigure}{.49\textwidth}
        \centering
        \includegraphics[width = \linewidth]{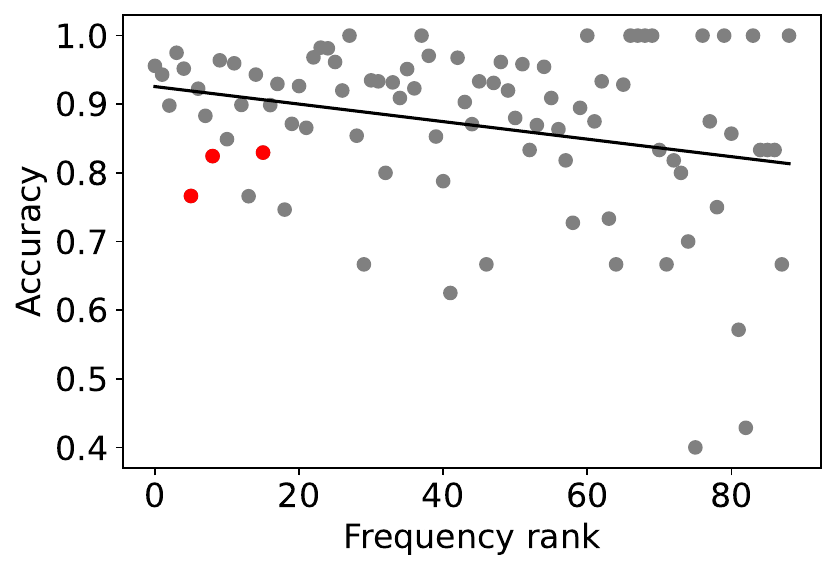}
        \caption{All chapters}
    \end{subfigure}\hfill
    \begin{subfigure}{.49\textwidth}
        \centering
        \includegraphics[width = \linewidth]{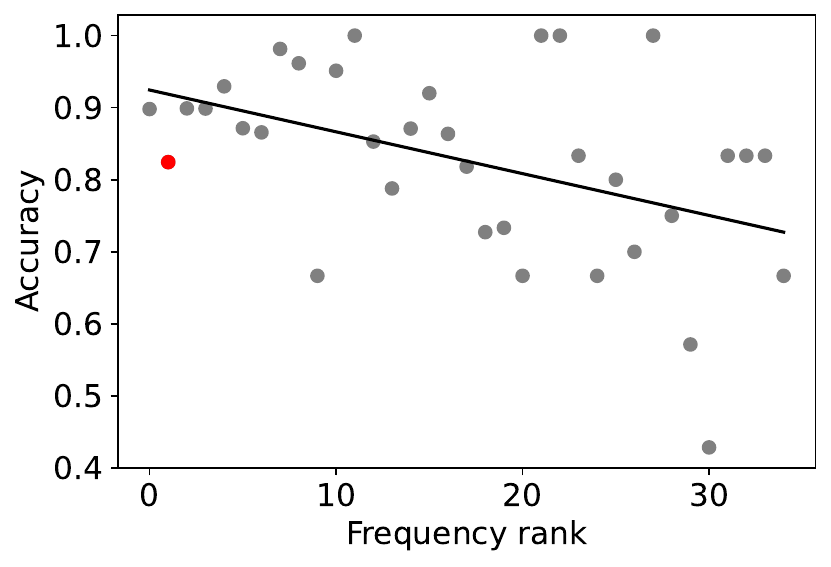}
        \caption{Chapter 84 (mechanical appliances)}
    \end{subfigure}\hfill
    \begin{subfigure}{.49\textwidth}
        \centering
        \includegraphics[width = \linewidth]{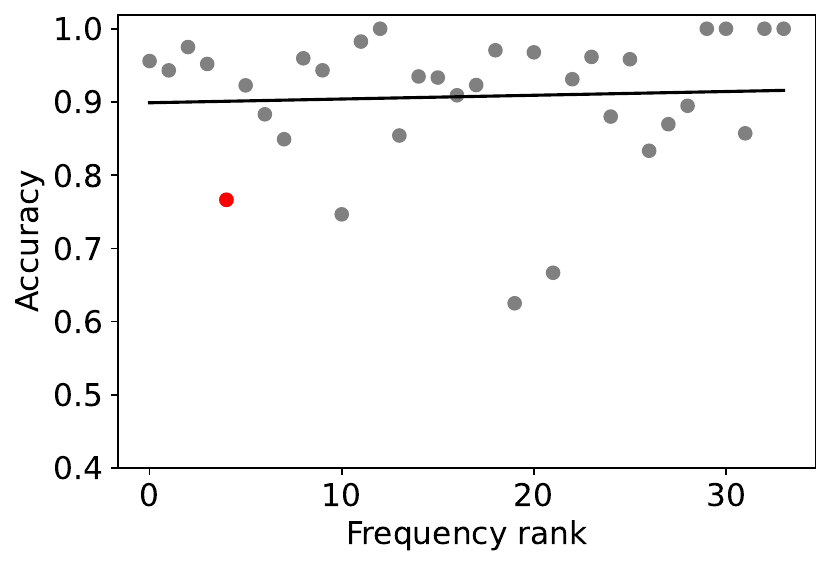}
        \caption{Chapter 85 (electronic equipment)}
    \end{subfigure}\hfill
    \begin{subfigure}{.49\textwidth}
        \centering
        \includegraphics[width = \linewidth]{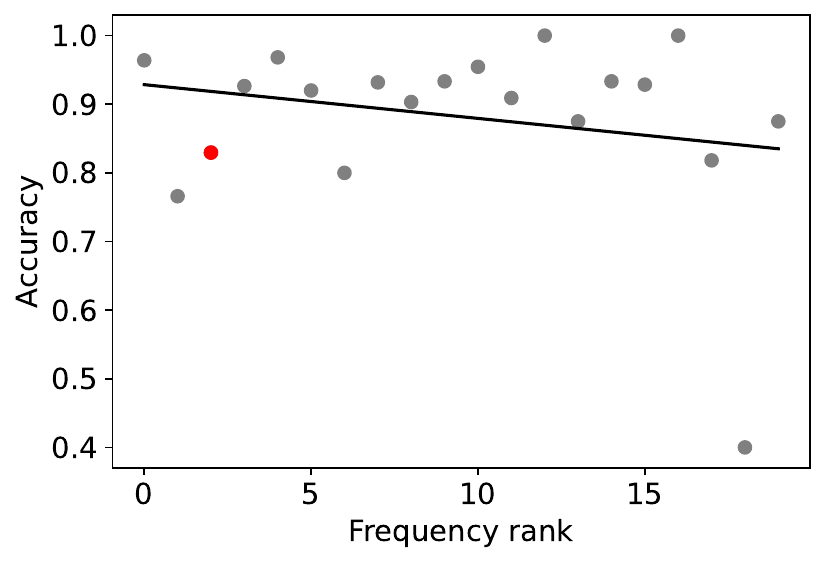}
        \caption{Chapter 90 (optical and photographic)}
    \end{subfigure} 
      \caption{The correlation between heading-level frequency and the prediction accuracy of the ELECTRA-based AI model. The AI model provided higher prediction accuracy for more frequently appearing goods in (b) Chapter 84 and (d) Chapter 90. (c) Chapter 85 goods include challenging cases and do not show the same desirable pattern. The red dots for each Chapter indicate the \textit{Miscellaneous} category, which shows below-average performance.}
\label{fig:etc_acc}
\end{figure*}

\subsection{Relationship between Frequency and Accuracy}

Figure~\ref{fig:etc_acc} shows the accuracy of each heading from the ELECTRA model. This figure depicts the trend in prediction accuracy as a function of heading frequency in the dataset. The negative slope of the fitted lines in Chapters 84 and 90 indicates that the AI model provided higher prediction accuracy for more frequently appearing goods. This is a desirable trait, as top-ranked items take up a disproportionately larger part of the entire data. However, Chapter 85 does not show the same characteristic, likely because it contains challenging cases. 

\label{subsec:others}
Each Chapter contains a heading category named \textit{Miscellaneous} which is used to hold a variety of cases that do not perfectly fit any given heading within the Chapter. This Miscellaneous category was one of the frequently requested categories for classification, yet as one may expect, the AI model also gave below-average prediction accuracy.

%% file: tables/main_results.tex
\begin{table}[htpb!]
\caption{Classification accuracy for heading (HS4) and subheading (HS6). The result shows that the top-3 suggestions made with three language models have an accuracy of about 90\% in classifying 925 subheadings.
}
\label{table:main_results}
\centering
\small
\resizebox{0.8\linewidth}{!}{
\begin{tabular}{l|ccc|ccc}\toprule
 & \multicolumn{3}{c|}{HS4} & \multicolumn{3}{c}{HS6}\\
\cline{2-7}\rule{0pt}{2.5ex}Model / Top-$k$ accuracy  & $k=1$ & $k=3$ & $k=5$ & $k=1$ & $k=3$ & $k=5$ \\
\midrule
LSTM-based & 50.96 & 65.88 & 72.10 & 36.02 & 53.46 & 62.00  \\
KoBERT & 84.51 & 91.07 & 92.77  & 78.88 & 87.70 & 90.06  \\
KoELECTRA & 87.48 & 93.42 & 94.92 & 83.22 & 90.99 & 92.88 \\
KLUE-RoBERTa & \textbf{89.24} & \textbf{95.50} & \textbf{96.49} & \textbf{86.06} & \textbf{93.98} & \textbf{95.25} \\

\bottomrule
\end{tabular}}
\end{table}

%% file: tables/data_difficulty.tex
\begin{table}[htpb!]
\caption{Classification accuracy for contentious cases. Compared to the challenging cases escalated to the HS \textsf{Committee and Council}, the AI model shows outstanding performance for general HS inquiries. We note that obvious cases are not requested for classification, but only the ones that exporters and importers cannot resolve are submitted to the Institute, and are handled as a \textsf{General} case.
}
\label{table:data_difficulty}
\centering
\small
\resizebox{0.8\linewidth}{!}{
\begin{tabular}{l|ccc|ccc}\toprule
 & \multicolumn{3}{c|}{HS4} & \multicolumn{3}{c}{HS6}\\
\cline{2-7}\rule{0pt}{2.5ex}Dataset / Top-$k$ accuracy  & $k=1$ & $k=3$ & $k=5$ & $k=1$ & $k=3$ & $k=5$ \\
\midrule
Committee \& Council & 63.04 & 84.70 &  89.13 & 58.70 & 73.17 & 84.78  \\
General & 89.36 & 95.60  & 96.56 & 86.31 & 94.17 & 95.35\\

\bottomrule
\end{tabular}}
\end{table}

%% file: tables/supporting_facts.tex
\begin{table}[htpb!]
\centering
\small
\caption{Comparison of sentences written by experts and sentences retrieved by our model in a sample case. Three out of four sentences retrieved by our model were equivalent to the expert's evidence sentences. 
}
\label{tab:supporting_facts}
\scalebox{1}{%
\begin{tabular}{ p{13cm}} \toprule
    \textbf{Reasons for decision by experts} \\
         1. 84.72 Other office machines (for example, hectograph or stencil  duplicating machines, addressing machines, automatic banknote dispensers, coin-sorting machines, coin-counting or wrapping machines, pencil-sharpening machines, perforating or stapling machines).\\ 
         2. This heading covers all office machines not covered by the preceding two headings or more specifically by any other heading of the Nomenclature. \\
         3. The term ``office machines” is to be taken in a wide general sense to include all machines used in offices, shops, factories, workshops, schools, railway stations, hotels, etc., for doing ``office work” (i.e., work concerning the writing, recording, sorting, filing, etc., of correspondence, documents, forms, records, accounts, etc.). \\
         4. Office machines are classified here only if they have a base for fixing or for placing on a table, desk, etc. The heading does not cover the hand tools, not having such a base, of Chapter 82. \\\midrule
    \textbf{Supporting facts found by our model} \\
         1. This heading covers all office machines not covered by the preceding two headings or more specifically by any other heading of the Nomenclature.   $\rightarrow$ \textbf{Eqv. to (2)} \\
         2. The term ``office machines” is to be taken in a wide general sense to include all machines used in offices, shops, factories, workshops, schools, railway stations, hotels, etc., for doing ``office work” (i.e., work concerning the writing, recording, sorting, filing, etc., of correspondence, documents, forms, records, accounts, etc.) $\rightarrow$ \textbf{Eqv. to (3)} \\
         3. Office machines are classified here only if they have a base for fixing or for placing on a table, desk, etc. The heading does not cover the hand tools, not having such a base, of Chapter 82. $\rightarrow$ \textbf{Eqv. to (4) }\\
         4. Automatic banknote dispensers, operating in conjunction with an automatic data processing machine, whether on line or off line. \\     
         \bottomrule
\end{tabular}
}
\end{table}

%% file: 6_qualitative_study.tex
\section{Qualitative Evaluation}

We next tested the usability of the AI model via a survey. To determine how field officers perceived an AI assistant tool, we built a prototype service that could be accessed via the Web. The prototype system did not require any understanding of the model's inner workings. Our collaborating partners at the Korea Customs Service (KCS) helped recruit field officers at the Korea Customs Valuation and Classification Institute (Number of participants, N=32), who tested the efficacy of the prototype AI model. Customs field officers with varying career experiences participated in our usability survey. Below, we describe key findings and feedback from this survey study.

\begin{figure*}[htpb!]
\centerline{
      \includegraphics[width=1\linewidth]{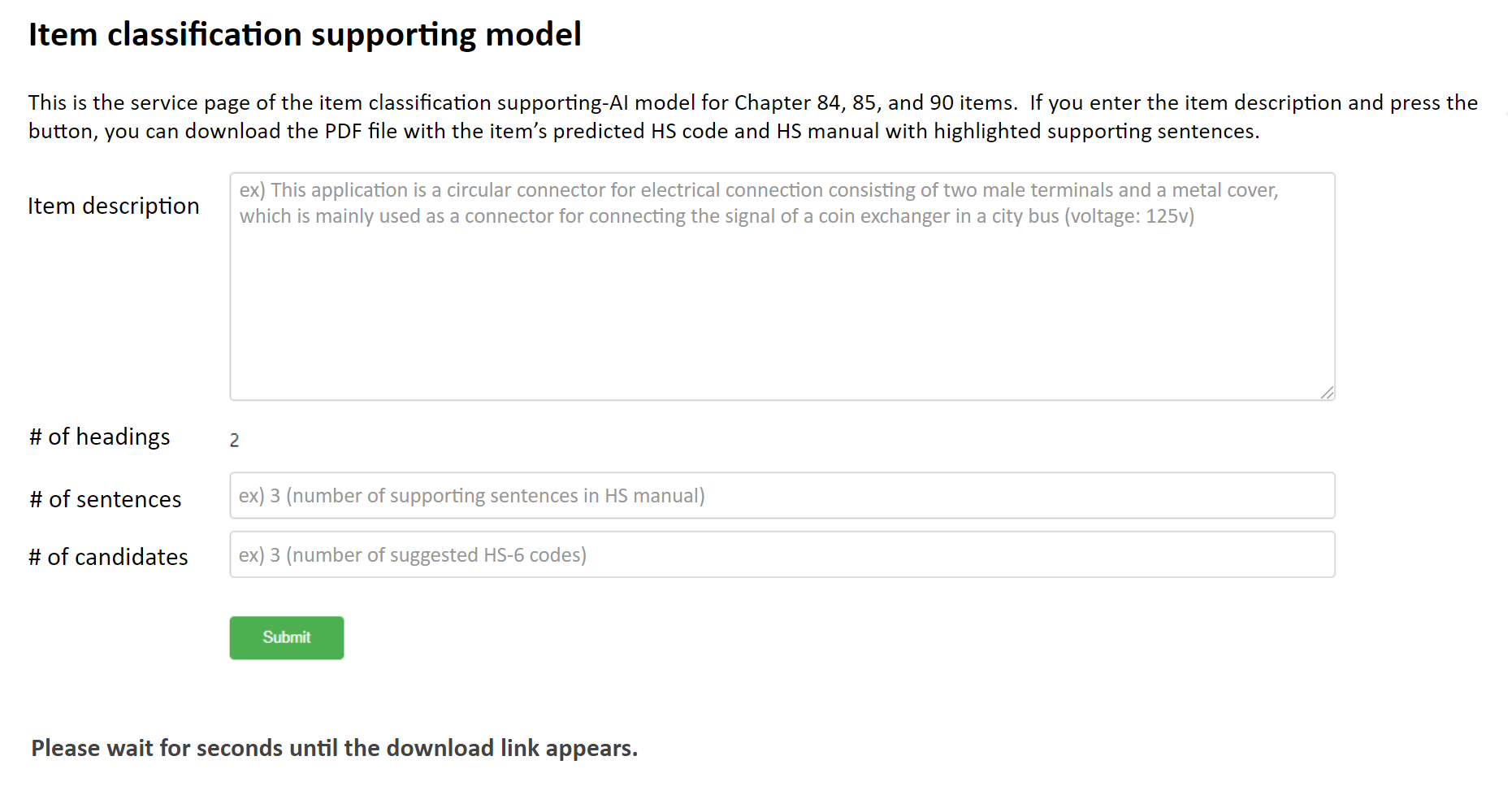}}
      \caption{Web interface of the prototype AI model tested by field officers} 
\label{fig:web_page}
\end{figure*}

\begin{figure*}[htpb!]
\centerline{
      \includegraphics[width=0.9\linewidth]{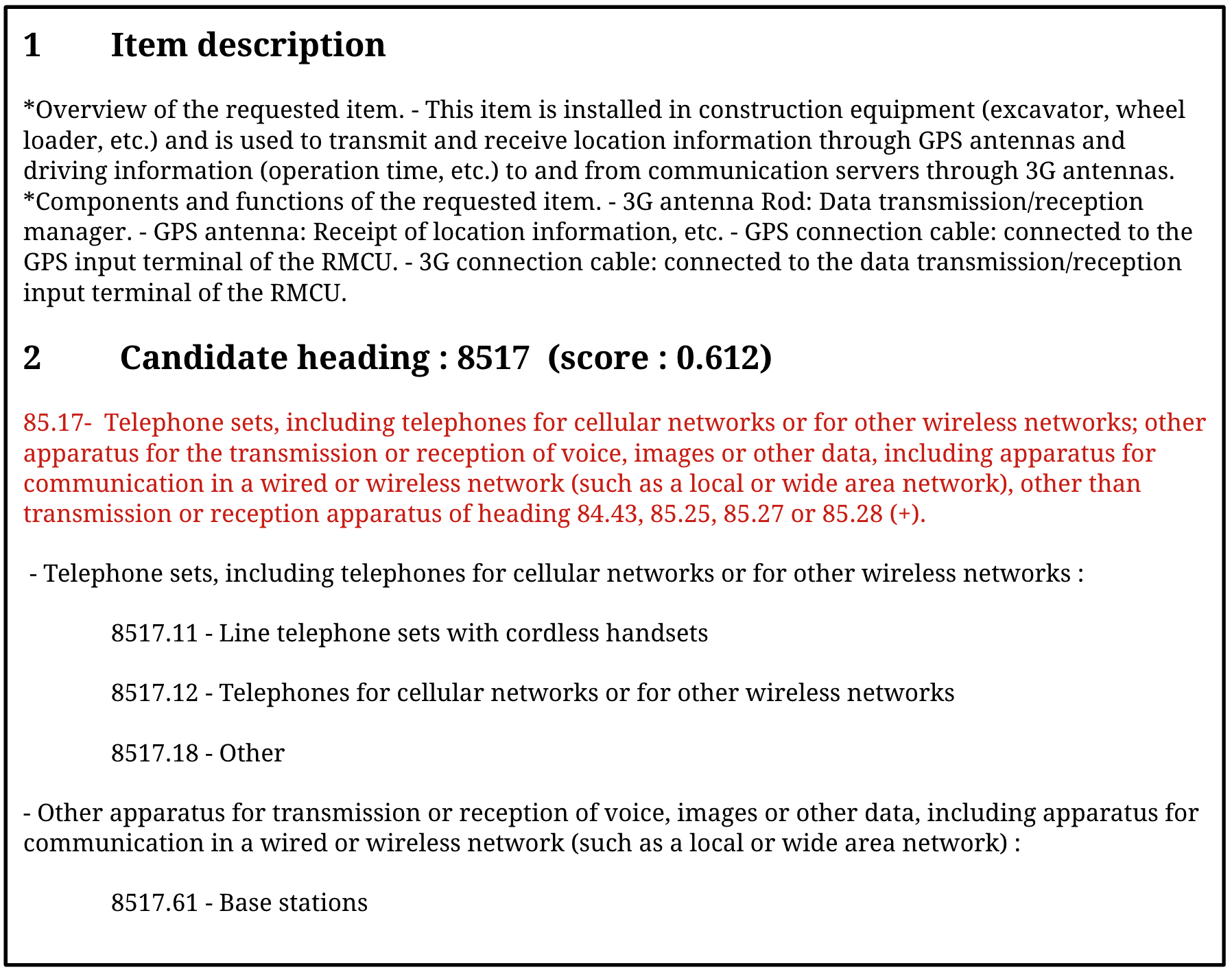}}
\centerline{
      \includegraphics[width=0.1\linewidth]{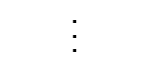}
      }
\centerline{
      \includegraphics[width=0.9\linewidth]{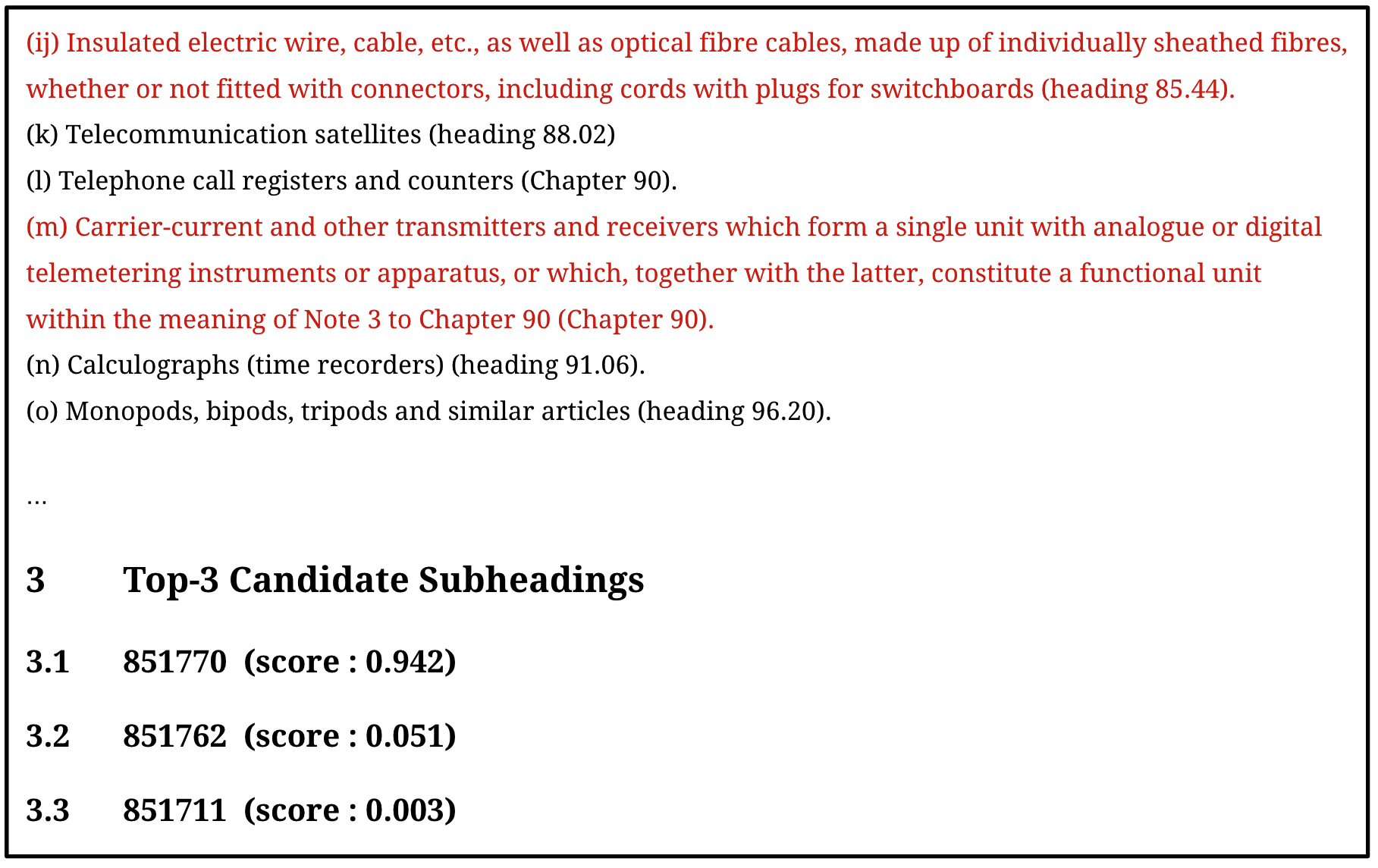}}
      \caption{Prototype model results provided to field officers. The entered item description, predicted heading, candidate heading's HS manual with supporting sentences, and candidate codes (subheading) are given.
      } 
\label{fig:output}
\end{figure*}

\subsection{Prototype Model}

Figure~\ref{fig:web_page} is a screen snapshot of the prototype Web service, which was designed to assist field officers in their daily classification tasks. The page allowed easy adjustment of the number of candidates and the length of the evidence sentences to be shown on the screen. Officers could type in or copy text descriptions of the goods they needed to classify, to start a new query. The top suggestions and the corresponding evidential sentences were given in a PDF file.

Figure~\ref{fig:output} is an example output. The output document consists of three parts: 1) entered item description, 2) candidate 4-digit heading with supporting sentences from the HS manual, 3) candidate 6-digit subheadings. The document includes the entire text of the candidate headings' HS manual, highlighting important sentences. We designed the output document with two goals in mind. The first was to mimic the HS Council and HS Committee's consulting documents which are produced for disputed cases. Like those documents, we quoted sentences from the HS manual that were the most characteristic of the suggestion. The second goal was to improve convenience for the field officers. Typically, the officers would refer to the HS manual if suggestions were relevant. Hence, we provided a complete version of the corresponding HS manual and highlighted the evidential sentences in red text. We also provided a calibrated prediction score for each candidate to indicate model confidence. Temperature scaling~\cite{guo2017calibration} was applied on $\mathbf{\hat{y}_i}$ to adjust the range of values.

\subsection{Qualification Analysis}

Our partners at the Korea Customs Service (KCS) helped recruit field officers of varying work experience to test the prototype. Participants were given two weeks to experience the prototype web page and received suggestions using the AI model in real-world situations. The usability survey was designed to evaluate the quality of the model's classification support under real-world situations, and 32 officers who tested the service responded to the survey. The survey included the following questions, including both Likert-scale and open-ended ones. Survey responses were anonymous: 
\begin{enumerate}
    \item How helpful was the AI suggestion? (Likert-scale was 1: not very helpful, 2: not helpful, 3: neutral, 4: helpful, 5: very helpful.) 
    \item Please describe if the assistant tool was helpful for your task, and if so, how?
    \item How accurate was the AI suggestion between 1 and 5? (Likert-scale was 1: not very accurate, 2: not accurate, 3: neutral, 4: accurate, 5: very accurate.)
    \item Please describe your thoughts on the accuracy of the AI suggestions.
    \item How many candidate suggestions and evidence sentences would you like to see?
    \item How long have you worked in the customs field and on the HS code classification task? \\
\end{enumerate}

\begin{figure*}[h]
    \begin{subfigure}{.49\textwidth}
        \centering
        \includegraphics[width = \linewidth]{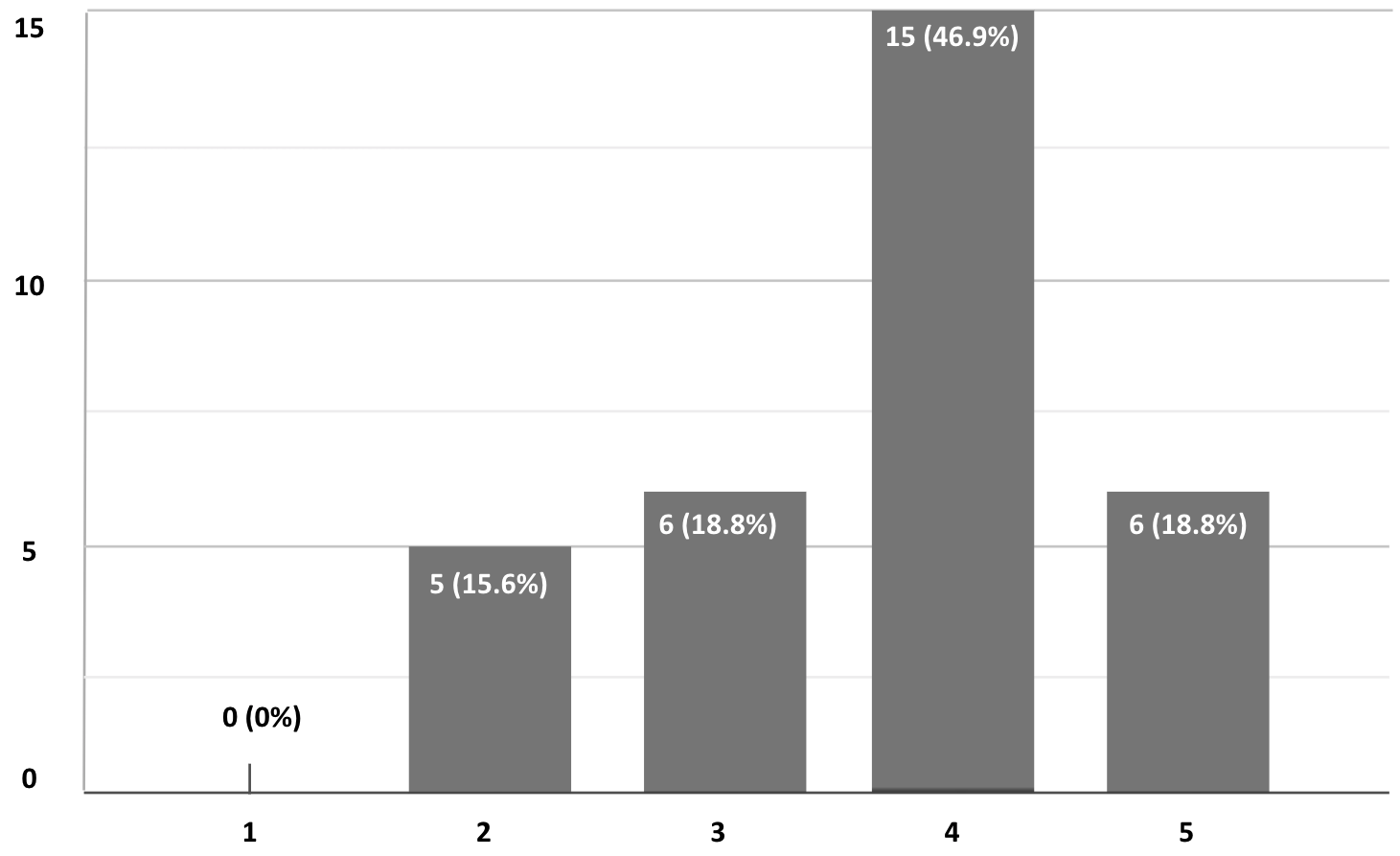}
        \caption{Helpfulness}
    \end{subfigure}\hfill
    \begin{subfigure}{.49\textwidth}
        \centering
        \includegraphics[width = \linewidth]{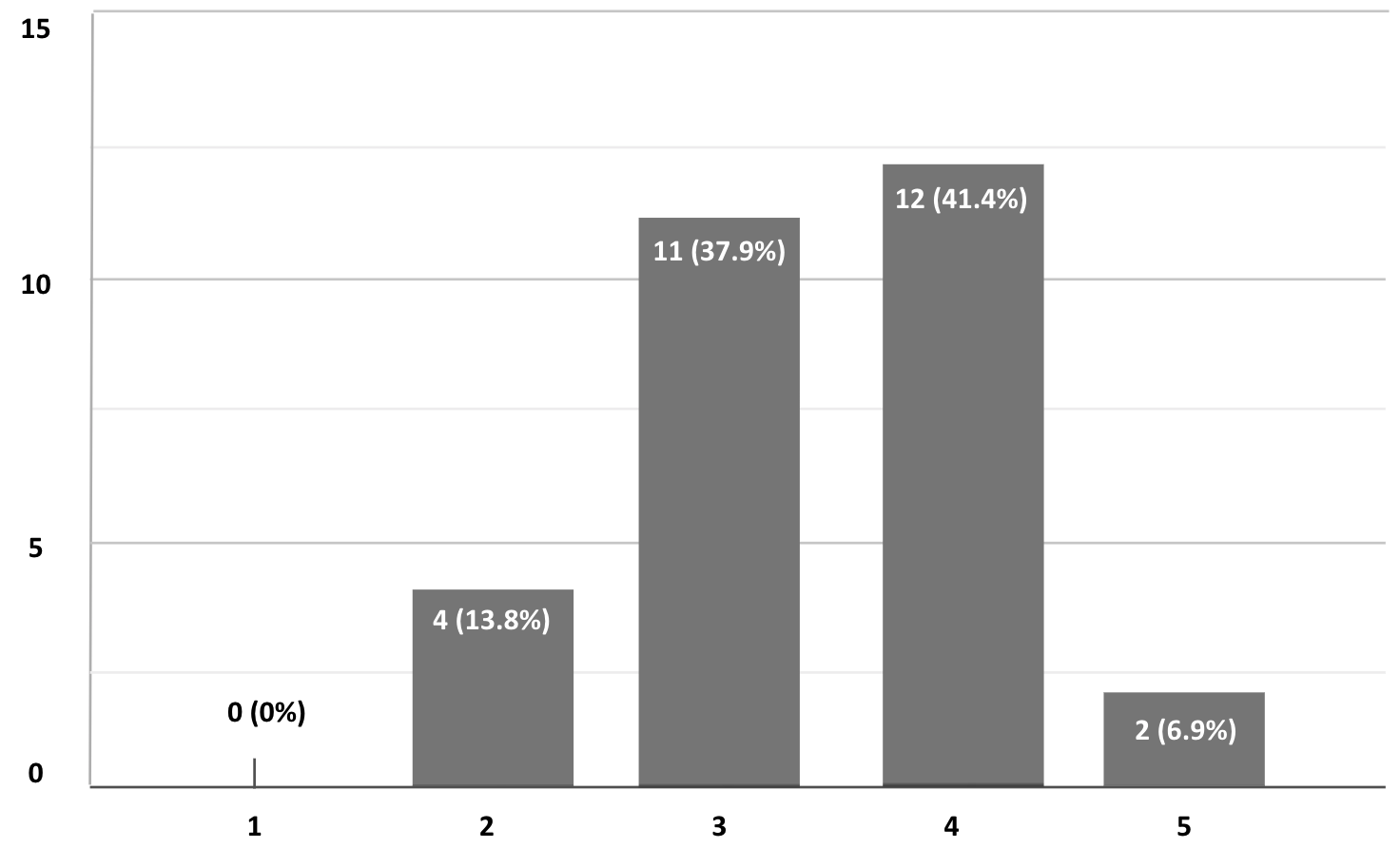}
        \caption{Accuracy}
    \end{subfigure}\hfill
      \caption{Results of the usability survey. (a) Distribution of helpfulness between 1 (not very helpful) and 5 (very helpful). (b) Distribution of accuracy between 1 (not very accurate) and 5 (very accurate). 
      } 
\label{fig:availability_accuracy}
\end{figure*}

\noindent $\bullet$ \textbf{Helpfulness and Accuracy. \quad}
Survey participants queried the system, with distinct goods and numerous variations in the number of candidate suggestions and evidential sentences as depicted in Figure~\ref{fig:web_page}. Figure~\ref{fig:availability_accuracy} shows the distribution of the helpfulness and accuracy responses.

65.7\% of the participants answered that the AI suggestions were helpful, with a score of four or five. Accuracy also showed a positive response overall. We found that more than 85\% gave a score of three or more for accuracy. In addition, there was a tendency that the participant who responded as `helpful' also rated higher accuracy. The Pearson correlation between the helpfulness score and accuracy was 0.82, and the Spearman correlation was 0.79. 

\noindent $\bullet$ \textbf{Analysis by Career Experience. \quad} 
Based on the final survey question about the career experience at customs and the HS classification task, we revisited the responses on helpfulness and accuracy. We identified five respondents each who had the longest work experience at customs as \textsf{Group A}, the shortest work experience at customs as \textsf{Group B}, the longest work experience at HS classification task as \textsf{Group C}, and the shortest work experience at HS classification task as \textsf{Group D}. The respondents in \textsf{Group A} on average had spent over 22 years in customs service, and \textsf{Group B} spent fewer than six years. \textsf{Group C} had worked on average longer than seven years on the classification task, and \textsf{Group D} had spent less than a year on the task.



\begin{figure*}[h]
    \begin{subfigure}{.49\textwidth}
        \centering
        \includegraphics[width = \linewidth]{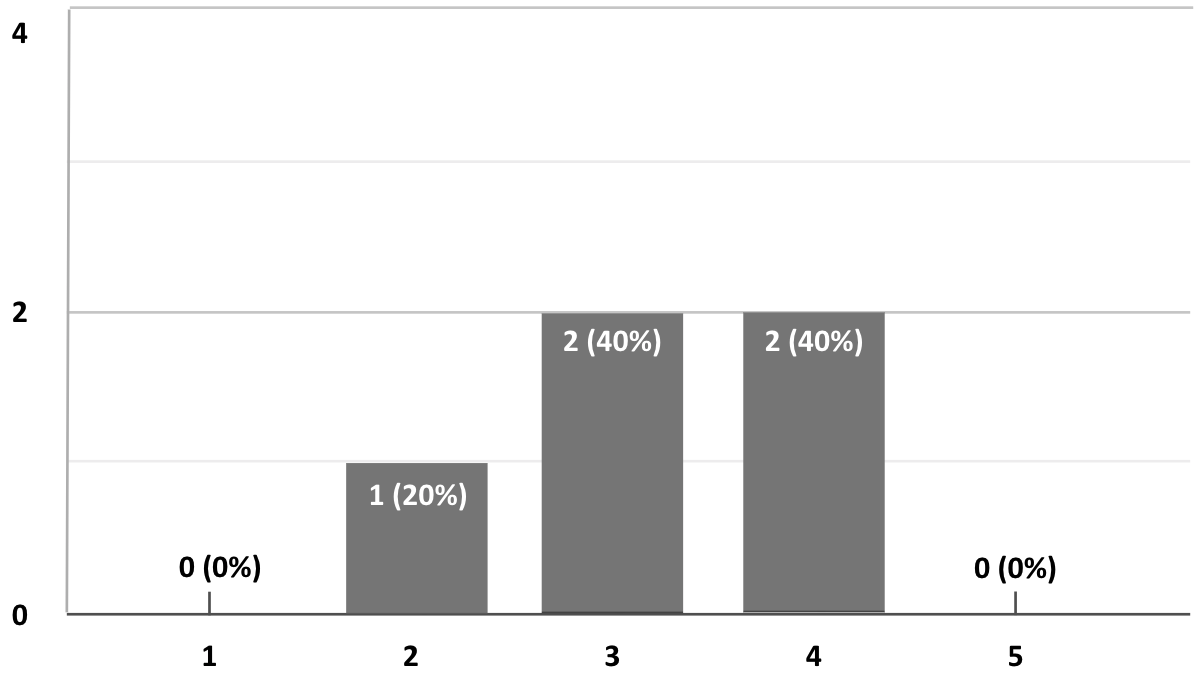}
        \caption{\textsf{Group A} (Experienced at customs)}
    \end{subfigure}\hfill
    \begin{subfigure}{.5\textwidth}
        \centering
        \includegraphics[width = \linewidth]{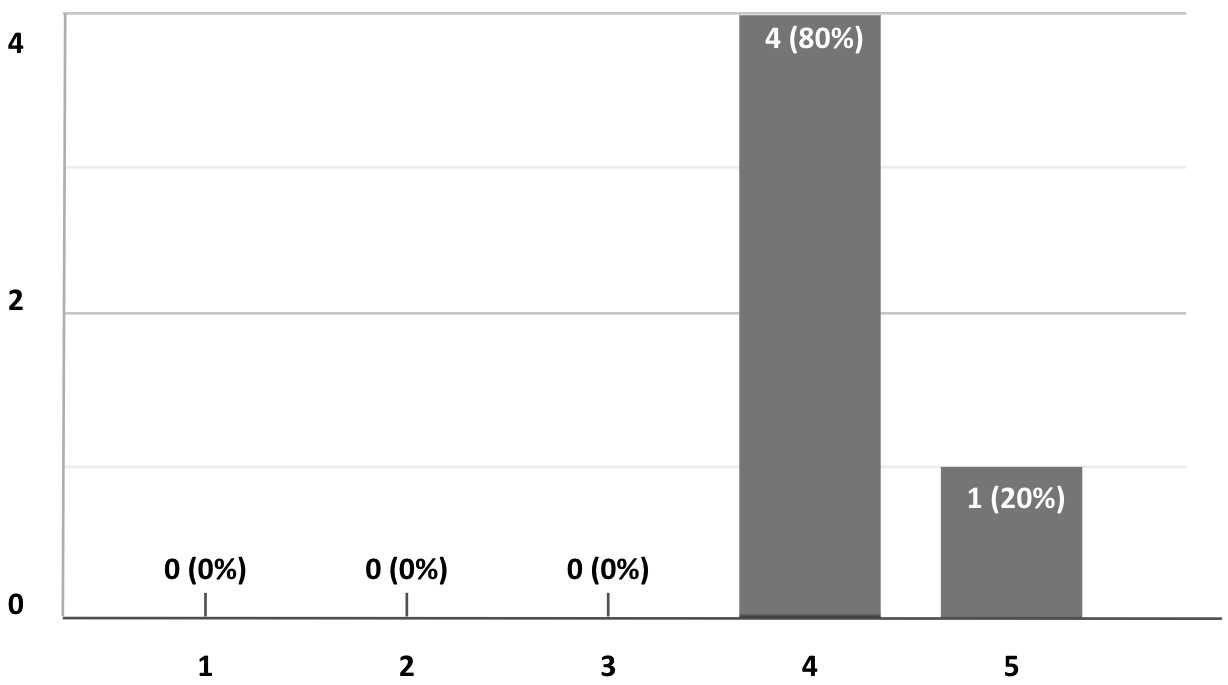}
        \caption{\textsf{Group B} (Less experienced at customs)}
    \end{subfigure}\hfill
    \begin{subfigure}{.49\textwidth}
        \centering
        \includegraphics[width = \linewidth]{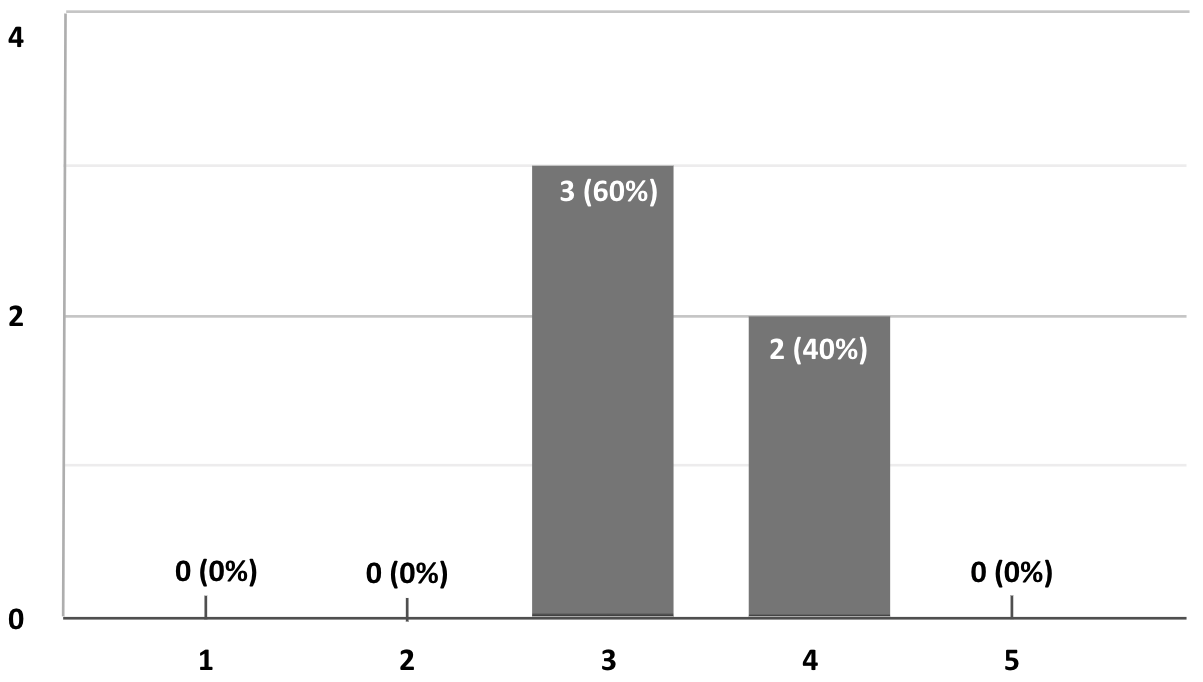}
        \caption{\textsf{Group C} (Experienced at classification)}
    \end{subfigure}\hfill
    \begin{subfigure}{.5\textwidth}
        \centering
        \includegraphics[width = \linewidth]{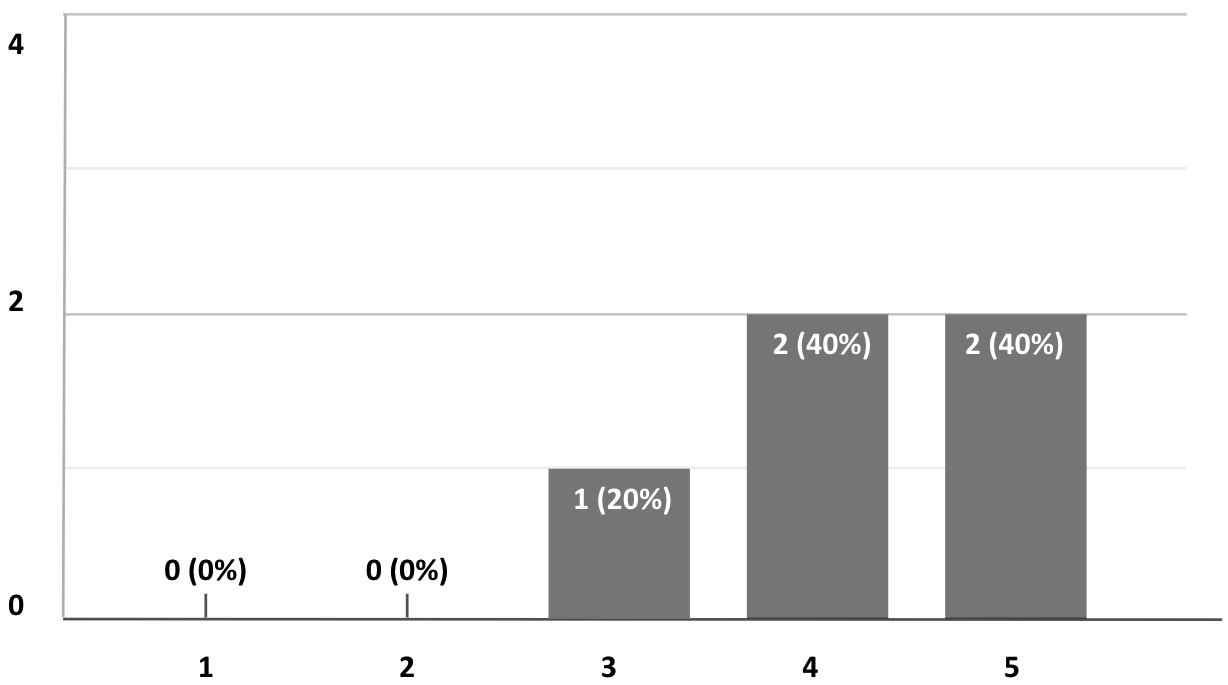}
        \caption{\textsf{Group D} (New at classification)}
    \end{subfigure}\hfill
      \caption{Distribution of helpfulness scores by work experience. (a) \textsf{Group A} has the longest work experience at customs service,  (b) \textsf{Group B} has the shortest work experience at customs service, (c) \textsf{Group C} has the longest work experience at classification task, and (d) \textsf{Group D} has the shortest work experience at classification task. 
      } 
\label{fig:group_result}
\end{figure*}

Figure~\ref{fig:group_result} shows the score of helpfulness answered by four groups. The top plots comparing \textsf{Group A} and \textsf{Group B} indicate that less experienced officers with a shorter working period in customs service found the AI assistant tool substantially more helpful. Four out of five respondents gave a score of 4, and one gave a score of 5. Even among the experienced officers with an average of 22 years in the career, two out of five respondents gave a score of 4 in helpfulness. 

The bottom two plots comparing \textsf{Group C} and \textsf{Group D} reinforced our finding that less experienced officers tended to find the AI tool more helpful. Among the least experienced officers who had been on the classification task for less than a year, two out of five gave a helpfulness score of 5.

\textcolor{black} {These findings suggest that AI models can effectively support novice customs officers in their tasks. The feedback received from senior officers indicates a desire for more precise and comprehensive information beyond what they already know from the system, resulting in relatively lower scores. This can be interpreted as an indication that the system possesses similar knowledge and classification abilities as the experienced officers themselves. } \\




\noindent $\bullet$ \textbf{Open-Ended Feedback. \quad} 
Next, we examined the open-ended feedback to understand what aspect of the AI tool was considered the most (or the least) helpful (or accurate). The overall feedback was positive, and some respondents were surprised with the level of accuracy the model could achieve on the challenging task. Here are some quotes from the feedback. 

One respondent shared a perspective that the AI tool acted as a `second eye' for decision making: 
\begin{quote}
``\textit{AI suggestions were helpful because I could compare my decision with them.}'' (\textsf{P1})
\end{quote}
Other participants also resonated with this perspective, since validation is critical in HS classification, knowing the decision directly affects the tariff rates and has a huge financial impact on importers and exporters. Field officers valued that AI could act as a validation tool for their own decisions. Similar to increasing the accuracy of classification, there was also a mention that the tool helped reduce potential errors.
\begin{quote}
``\textit{The model gave suggestions that I could have missed. I found this very helpful.}'' (\textsf{P13})
\end{quote}


Another common perspective shared was the tool's ability to reduce the time needed for initial investigation. Many field officers started by listing a wide pool of candidate HS codes, and then filtering down to a smaller set of candidate decisions. The AI tool was able to assist this initial investigation by introducing a large number of candidate suggestions. Here are some relevant quotes:  
\begin{quote}
``\textit{I think this tool can help shorten my screening time.}'' (\textsf{P14})
\end{quote}
\begin{quote}
``\textit{The model helped me make quick decisions.}'' (\textsf{P29})
\end{quote}

When it came to helpfulness, several participants responded that candidate codes and the evidential sentences highlighted from the HS manual were useful in making a decision:
\begin{quote}
``\textit{The algorithmic suggestions and the supporting sentences helped me reduce the candidate pool to review.}'' (\textsf{P7})
\end{quote}
\begin{quote}
``\textit{I found the snippet of HS manual given with candidate suggestions helpful. I could concentrate on the model's reasoning sentences and references.}'' (\textsf{P3})
\end{quote} 

Some officers mentioned the potential for the tool to be used as an educational tool, as it could give an overview of the classification task. The feedback here could also be appreciated together with the high accuracy and helpfulness scores shown for the less experienced officers. Here are some relevant quotes:
\begin{quote}
``\textit{The supporting model gave a rough idea of final decision I had to make.}'' (\textsf{P20})
\end{quote}
\begin{quote}
``\textit{Since the model shows the candidates, it can be helpful to educate new workers who have short working experience and expertise in the classification task.}'' (\textsf{P12})
\end{quote}

Last, related to how many candidates and supporting evidence individuals wanted to see, participants were most comfortable with the visual setting when the AI model provided three candidate subheadings and seven evidential sentences. The exact average was 3.17 subheadings and 6.74 sentences each. \textcolor{black} {Notably, when examining the preferences of junior officers (the 10 least experienced officers) and senior officers (the 10 most experienced officers), it was observed that junior officers favored 3.0 candidates and 5.8 sentences, while senior officers leaned towards 3.5 candidates and 5.4 sentences. However, despite these variations, the differences in responses between the two groups were not substantial. It is noteworthy that both junior and senior officers exhibited a similar viewpoint regarding the desired number of candidates and sentences.}

%% file: 7_discussion.tex
\section{Discussion}

\subsection{Toward Interpretable Results}
Interpretability is critical in many high-risk scenarios like HS classification. We provided a confidence score for each HS code candidate, which provides additional information to help the user judge whether or not a candidate was valid.
Although the score is tuned by temperature scaling, the range of the top-$k$ confidence score is quite different for each input item. Careful calibration is required to encourage customs officers to use this value as a reference in decision-making.


Another way to increase interpretability is to visualize the part of the item description related to each subheading candidate; then, customs officers can concentrate on the selected part and decide whether to consider second and third candidates to review. In addition, key sentences should relate to the subheading characteristics so that the final form of model output resembles the reports written by experts. Creating an organized document that explains the relations among prediction, description, and HS manual can reduce the effort required for HS code classification.

These records pertained to the resolution
of contentious cases undertaken by the HS Committee and HS Council


As this work demonstrated, there is a substantial resemblance in how customs experts decide on the HS codes with how judges decide on legal cases~\cite{winter2021judicial}. Moreover, deep learning models solve classification problems by finding common patterns from previous cases. As a result, past examples are the primary determinants of the AI model's decision, unlike human experts, who make decisions based on rules and manuals. Since the HS code and its manual undergo revisions every five years, previous cases cannot always be good references for recent ones. Revision can be viewed as an update of the knowledge base. This makes it essential to employ GRIs and the HS manual to prepare a credible model, which utilizes contextual information in model training based on deep linguistic understanding~\cite{wiegreffe2021explainableNLP}.

\subsection{Challenges}
\begin{figure*}[htpb!]
    \begin{subfigure}{.49\textwidth}
        \centering
        \includegraphics[width = \linewidth]{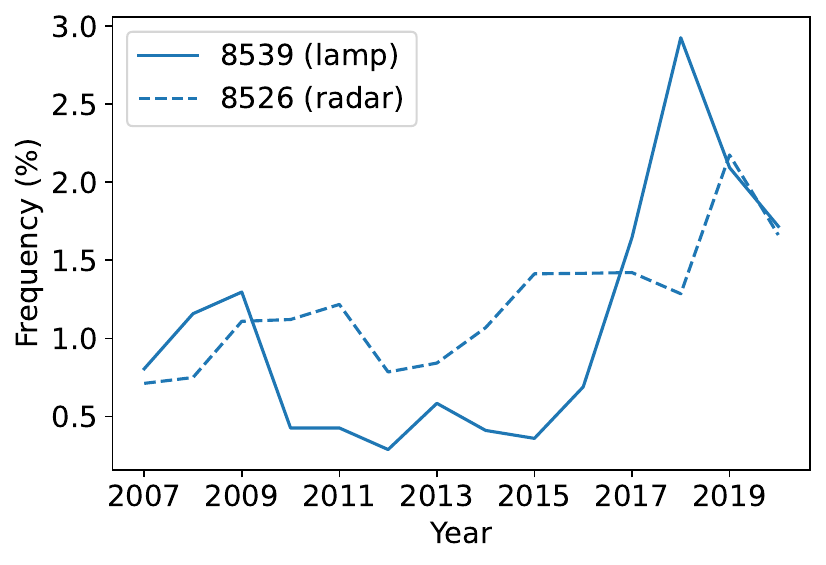}
        \caption{Headings whose share increased over time. Heading 8539 represents sealed beam lamp units and arc lamps, and 8526 is for radar apparatus or radio remote control apparatus including GPS tracker.}
    \end{subfigure}\hfill
    \begin{subfigure}{.49\textwidth}
        \centering
        \includegraphics[width = \linewidth]{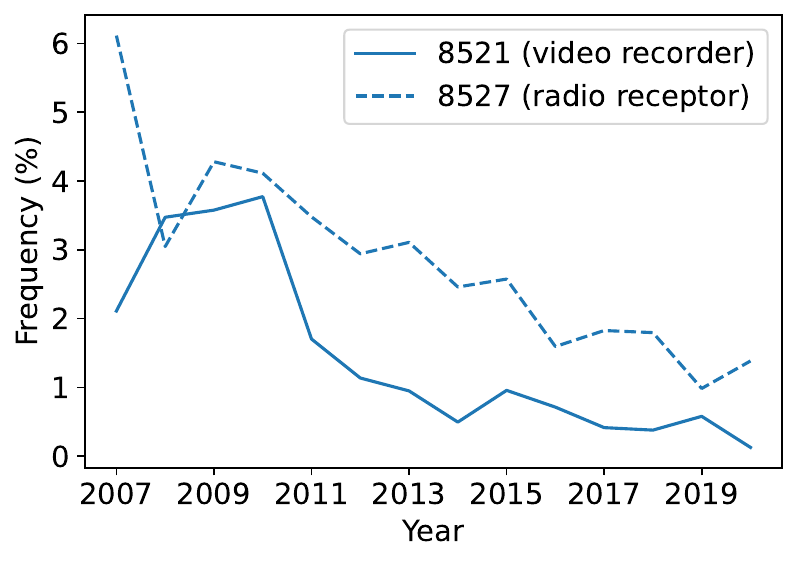}
        \caption{Headings whose share decreased over time. Heading 8521 includes video recording or reproducing apparatus, and 8527 is for reception apparatus for radio-broadcasting.}
    \end{subfigure}
    \caption{Data distribution shift: Four representative headings show changes in trade patterns over time.}
\label{fig:data_time}
\end{figure*}

\noindent $\bullet$ \textbf{Data Distribution Shift. \quad}
As the AI model is based on the learned knowledge from existing classification, a shift in data distribution can affect the quality of the suggestion. However, such shifts are a common part of many live systems, and this poses a challenge for the longitudinal adoption of an AI assistant model. Trade data is no exception. Figure~\ref{fig:data_time} shows example headings that have been increasing in their request at the HS classification task, at the same time other requests have been decreasing over a decade. This natural change reflects the adoption of new technology in developing products. One may re-learn these changes by updating the AI model's knowledge from time to time or by attempting to continue learning new patterns.


Another aspect is an update in the HS manual itself, which goes through extensive revision and updates every five years to represent technological advances better. For example, when the smartwatch was launched, there was only an HS code for smart mobile devices and a traditional watch, but not a combination. Choosing either selection will lead to different tariff rates. The same goes for many other technological advances. The World Customs Organization (WCO) considers these changes and introduces new subheadings or removes subheadings that are no longer used, etc. The most recent release of the HS manual is the seventh edition which had 351 sets of amendments, effective from January 1, 2022\footnote{http://www.wcoomd.org/en/topics/nomenclature/instrument-and-tools/hs-nomenclature-2022-edition.aspx}. In this new edition, subheading \textit{8517.13} representing `\textit{Smartphones}' newly appears. Previously, smartphones were usually classified into \textit{8517.12} `Telephones for cellular networks or other wireless networks.' 


As mentioned earlier, the AI model needs to take these new changes into account and re-learn the manual. Without a specific update, the model suggestions will no longer be relevant over time~\cite{gate}. Once a model has been deployed, continual monitoring of the parameter stability and model performance is required~\cite{huyen2022datashift}. The maintainer should retrain the model whenever the system detects a distribution shift. \textit{Continual learning} is a method that allows such model training on new data~\cite{mai2021drift} and this idea can be applied to our AI model as an extension. \\



\noindent $\bullet$ \textbf{Data Imbalance and Vagueness. \quad}
Another challenge in building the AI model is data imbalance and vagueness. As we have shown in Figure~\ref{fig:data_imbalance}, the headings and subheadings follow a skewed distribution. Some appear disproportionately more in the data, whereas others appear substantially less frequently. Together with the temporal nature of data, data imbalance could pose a challenge in predicting non-popular items.


\begin{figure*}[htpb!]
    \begin{subfigure}{.49\textwidth}
        \centering
        \includegraphics[width = \linewidth]{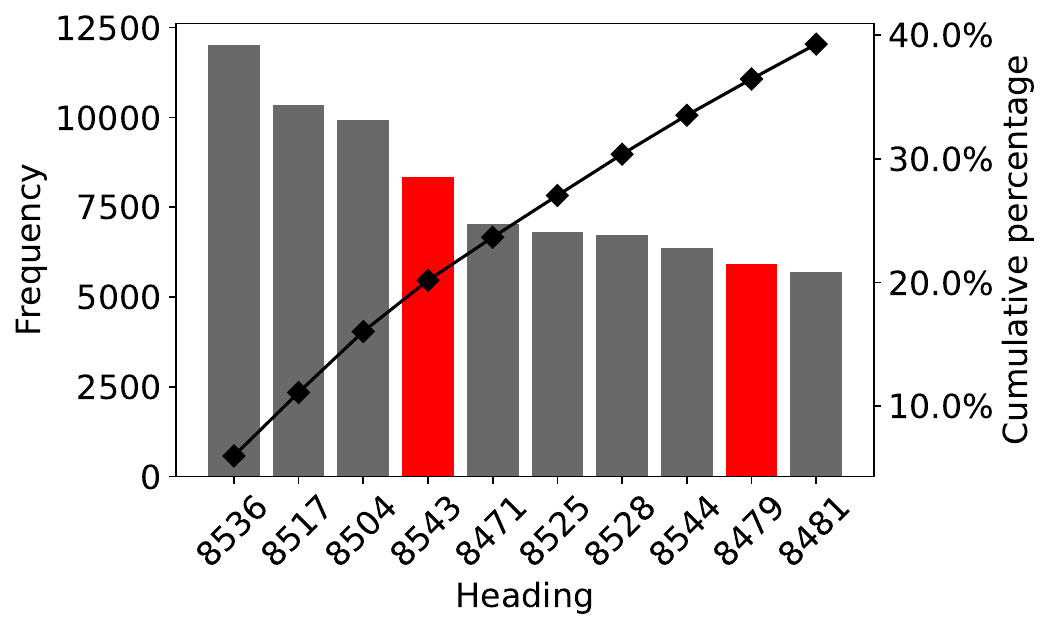}
        \caption{All chapters}
    \end{subfigure}\hfill
    \begin{subfigure}{.49\textwidth}
        \centering
        \includegraphics[width = \linewidth]{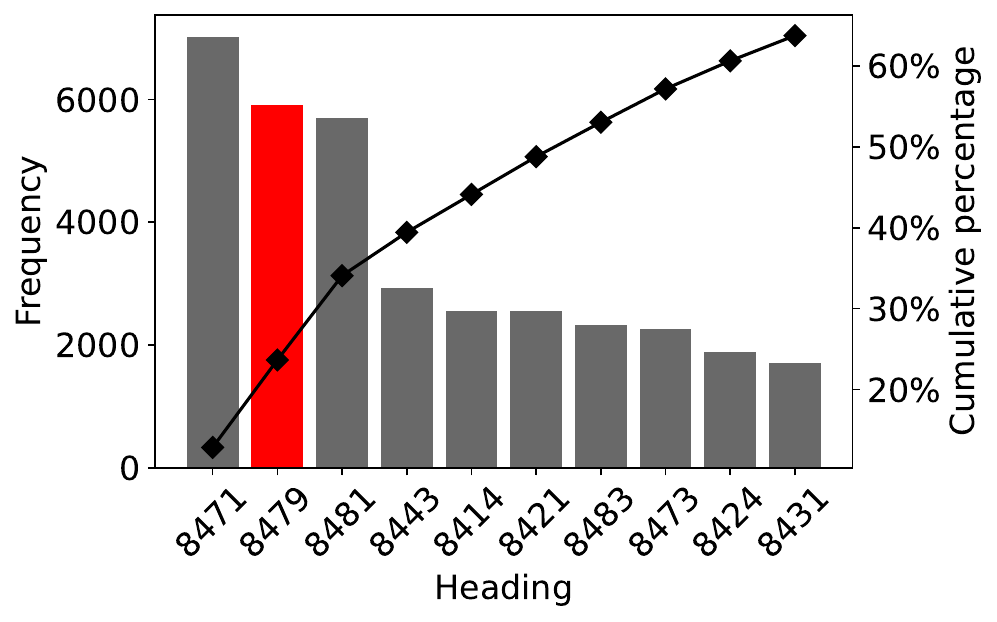}
        \caption{Chapter 84}
    \end{subfigure}\hfill
    \begin{subfigure}{.49\textwidth}
        \centering
        \includegraphics[width = \linewidth]{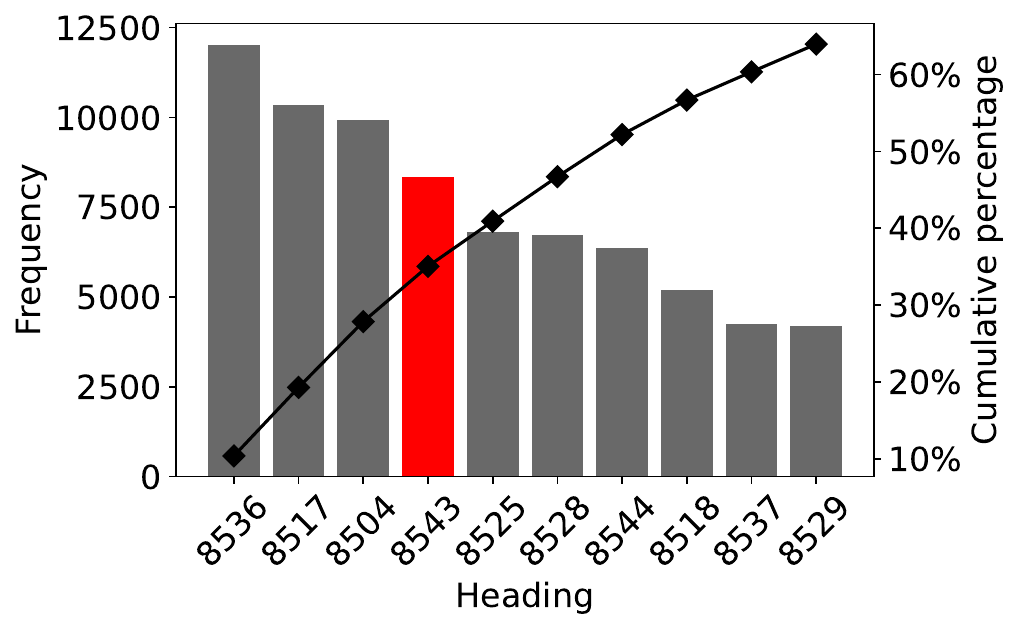}
        \caption{Chapter 85}
    \end{subfigure}\hfill
    \begin{subfigure}{.49\textwidth}
        \centering
        \includegraphics[width = \linewidth]{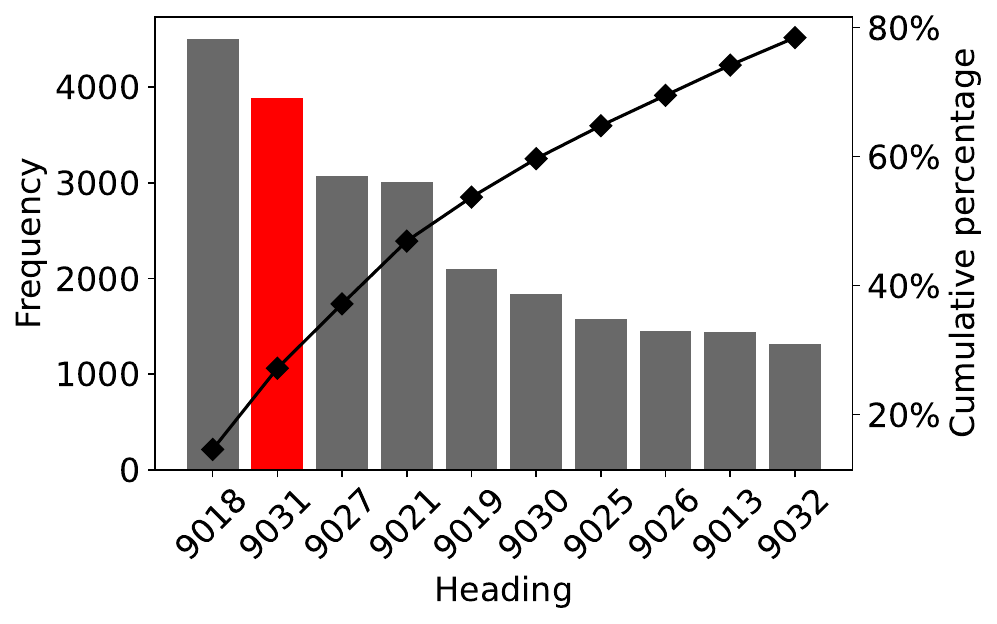}
        \caption{Chapter 90}
    \end{subfigure}
    \caption{Top-10 frequent headings in each chapter. Red bars indicate \textit{Miscellaneous} headings, which occupy a fairly large proportion.}
\label{fig:etc_f}
\end{figure*}

As the final challenge, we point back to the prevalence of the \textit{Miscellaneous} category. We have shown that this category is more difficult to predict than others, as it contains a variety of items that do not fit perfectly with other subheadings or headings. Figure~\ref{fig:etc_f} shows this challenge once more, indicating the popularity of the \textit{Miscellaneous} category (marked in red color) along with their proportions in the data. This category continues to become larger with new technological advances and will likely remain a challenging case to predict for both AI and humans. \\

\textcolor{black} {
\subsection{Possibility of Utilizing Large Language Models (LLMs)}}

\textcolor{black} {In order to explore the capabilities of the latest large language models, specifically ChatGPT, we conducted a series of toy experiments to determine whether LLM could classify customs products without the need for additional training steps. Our approach involved providing ChatGPT with the General Rules for the Interpretation of the Harmonized System (GRI) and a single heading-level HS manual. Specifically, we focused on the `8533: Electrical resistors' heading, which comprises seven subheadings as sub-classes. We presented ChatGPT with five item descriptions belonging to the `8533' chapter and tasked it with predicting the appropriate subheading for each item.\\
Remarkably, ChatGPT successfully predicted the correct subheading for four out of the five items. However, it encountered a misclassification issue, erroneously assigning the sub-heading `8533.31' to `8533.39' due to a misunderstanding of the numerical details within the item characteristics. Despite this classification test took place with the small number of class candidates, we observed that LLM can comprehend the standards associated with each HS code in a zero-shot manner. Based on these encouraging results from our toy experiment, we believe there is potential for applying the latest LLMs to address complex classification problems in this domain.}

\textcolor{black}{
Additionally, an intriguing aspect of ChatGPT's performance was its ability to provide explanations for its classification decisions, along with the predicted class, when the request was solely focused on classification. This interpretability feature holds promise for facilitating the handling of rapidly changing customs item classification tasks and generating understandable outcomes.}

\textcolor{black}{
Nevertheless, it is important to acknowledge the existing limitations. To further assess the capabilities of ChatGPT, we introduced two more HS manuals for the `8534' and `8535' classes, in total 14 subheadings. Subsequently, we provided ChatGPT with five item descriptions and requested classification tasks in conditions where there were more candidates to choose from. In this scenario, ChatGPT achieved three correct predictions out of five, and it also exhibited a common challenge: providing plausible yet incorrect answers. Notably, it misclassified the last product as `8536.69 - Electrical apparatus for connecting electrical circuits, for a voltage exceeding 1,000 V,' which is a non-existing subheading. Given the sensitivity of customs item classification, the issues of ambiguity and data insecurity pose significant challenges when leveraging trained LLMs. Addressing these concerns represents a crucial area for future research to harness the potential of these valuable tools effectively.}\\



%% file: 8_conclusion.tex
\section{Conclusions}

This research presents an AI model for assisting with the process of HS code classification at customs offices. Using the product description and the HS manual, the model predicted the first 4-digit headings and 6-digit subheadings along with supporting facts related to the suggestion to human experts. Rather than replacing human judgment entirely, the model gave top suggestions as a guiding tool. We also had a unique opportunity to collaborate with the Korea Customs Service and test the feasibility of the AI model as a prototype service with field officers (N=32).

We expect that our work will contribute substantially in various respects. The use of this AI assistant tool by declarants can improve the initial declaration quality, thereby reducing workload at customs offices, particularly when competing HS codes are problematic for declarants and customs officials. Internally, the tool could assist customs officials in the various ways identified in the survey, for example, as an educational tool for new officials, as a validation tool for experienced officials, and as a guiding tool that helps reduce the time and effort needed to screen for candidate codes by all officials. 

Our model presents the competing HS codes of the target product with its rationale, so it has great significance as an auxiliary means for product classification. Platforms require a systematic classification system to effectively expose and recommend products to users, but each product-providing company often has different standards. The platforms build and utilize hierarchical classification algorithms to maintain the consistent categorization of hundreds of millions of products. Our work can be used to advance these algorithms and classifications and facilitate their management. 

Looking to the future, large language models (LLMs) offer promising solutions for addressing the complexities of customs classification tasks. LLMs possess the ability to understand the nuances of customs item descriptions through their extensive pre-training. However, it's essential to cautiously integrate LLMs into our proposed algorithm while considering the risks associated with prediction errors, particularly hallucinations. In closing, our research has introduced a robust and transparent classification model for customs goods. Future work should prioritize the effective utilization of LLMs to enhance model adoption and performance within this context.